\documentclass{article} 
\usepackage{template/iclr2026_conference,times}

\usepackage{amsmath,amsfonts,bm}

\def\eqref#1{equation~\ref{#1}}

\def\1{\bm{1}}

\DeclareMathAlphabet{\mathsfit}{\encodingdefault}{\sfdefault}{m}{sl}
\SetMathAlphabet{\mathsfit}{bold}{\encodingdefault}{\sfdefault}{bx}{n}

\usepackage{hyperref}
\usepackage{url}

\usepackage{caption}
\usepackage{graphicx}
\usepackage{multirow}
\usepackage[table,xcdraw]{xcolor}
\usepackage{array}
\usepackage{booktabs}
\usepackage{enumitem}
\usepackage{makecell}
\usepackage{wrapfig}
\definecolor{oliveGreen}{HTML}{7A9A8D}
\definecolor{steelBlue}{HTML}{5B7C99}
\definecolor{mauve}{HTML}{B38BA4}
\hypersetup{
colorlinks,
linkcolor=oliveGreen,
anchorcolor=oliveGreen,
urlcolor=steelBlue,
citecolor=steelBlue,
}
\usepackage{pifont}
\newcommand{\cmark}{\ding{51}}

\usepackage{algorithm}
\usepackage[noend]{algpseudocode} 
\algrenewcommand\algorithmicrequire{\textbf{Input:}}
\algrenewcommand\algorithmicensure{\textbf{Output:}}
\algrenewcommand\textproc{\textsc} 
\algrenewcommand{\algorithmiccomment}[1]{\hfill\textcolor{oliveGreen}{// \textit{#1}}}
\algrenewcommand\alglinenumber[1]{\footnotesize #1}
\newcommand{\ID}[1]{\mathrm{ID}\!\left[#1\right]}
\newcommand{\tokThink}{\langle\mathrm{think}\rangle}
\newcommand{\tokEnd}{\langle/\mathrm{think}\rangle}
\usepackage{xspace}
\newcommand{\swireasoning}{\textsc{SwiReasoning}\xspace}

\usepackage{colortbl}
\usepackage{tikz}
\definecolor{MistyRose}{RGB}{255,228,225}  
\newcommand{\gaincolor}[1]{\textcolor{oliveGreen}{#1}}
\newcommand{\BARMAX}{5.0}

\newcommand{\Deltabar}[1]{%
\begin{tikzpicture}[baseline]
  \def\W{1.4}  
  \def\H{0.18}  
  \def\PAD{0.07}
  \draw[gray!40,line width=0.15pt] (0.5*\W,0) -- (0.5*\W,\H); 

  \pgfmathsetmacro{\len}{(\W/2.0)*((#1)/\BARMAX)}  
  \pgfmathsetmacro{\absd}{abs(#1)}
  \pgfmathsetmacro{\xzero}{0.5*\W}                 
  \pgfmathsetmacro{\xtip}{\xzero+\len}             

  \ifdim #1 pt>0pt
    \filldraw[oliveGreen]    (\xzero,0) rectangle (\xtip,\H);
    \pgfmathsetmacro{\xlbl}{min(\xtip+\PAD,\W)}
    \node[anchor=west,inner sep=0pt] at (\xlbl,0.5*\H)
      {\scriptsize \gaincolor{+\pgfmathprintnumber[fixed,precision=2,zerofill]{\absd}}};
  \else
    \filldraw[oliveGreen!30] (\xtip,0) rectangle (\xzero,\H);
    \pgfmathsetmacro{\xlbl}{max(\xtip-\PAD,0)}
    \node[anchor=east,inner sep=0pt] at (\xlbl,0.5*\H)
      {\scriptsize \gaincolor{-\pgfmathprintnumber[fixed,precision=2,zerofill]{\absd}}};
  \fi
\end{tikzpicture}%
}

\usepackage{textcomp}  
\usepackage{scalerel}  
\def\hf{\scalerel*{\includegraphics{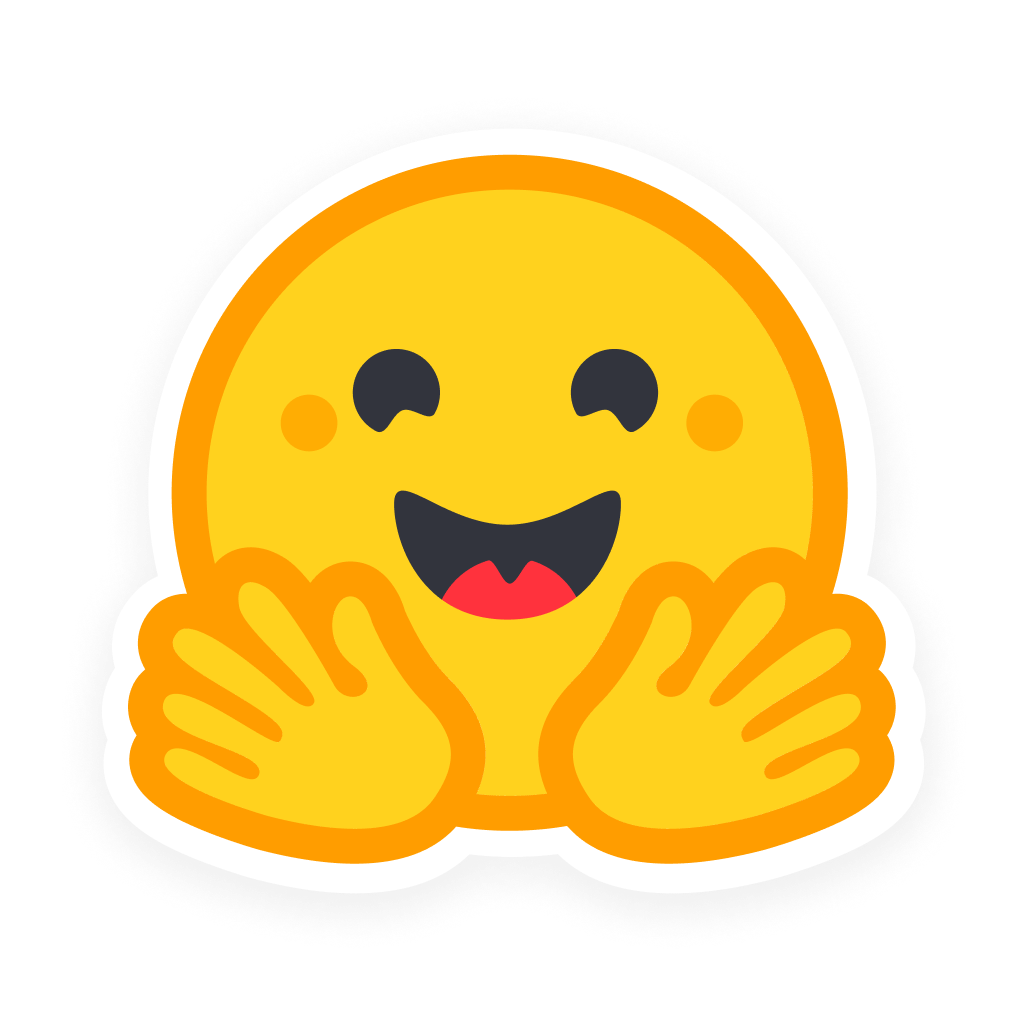}}{\textrm{\textbigcircle}}}
\def\github{\scalerel*{\includegraphics{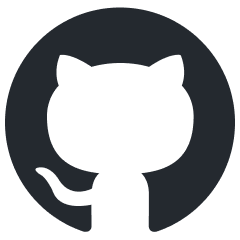}}{\textrm{\textbigcircle}}}
\def\webiste{\scalerel*{\includegraphics{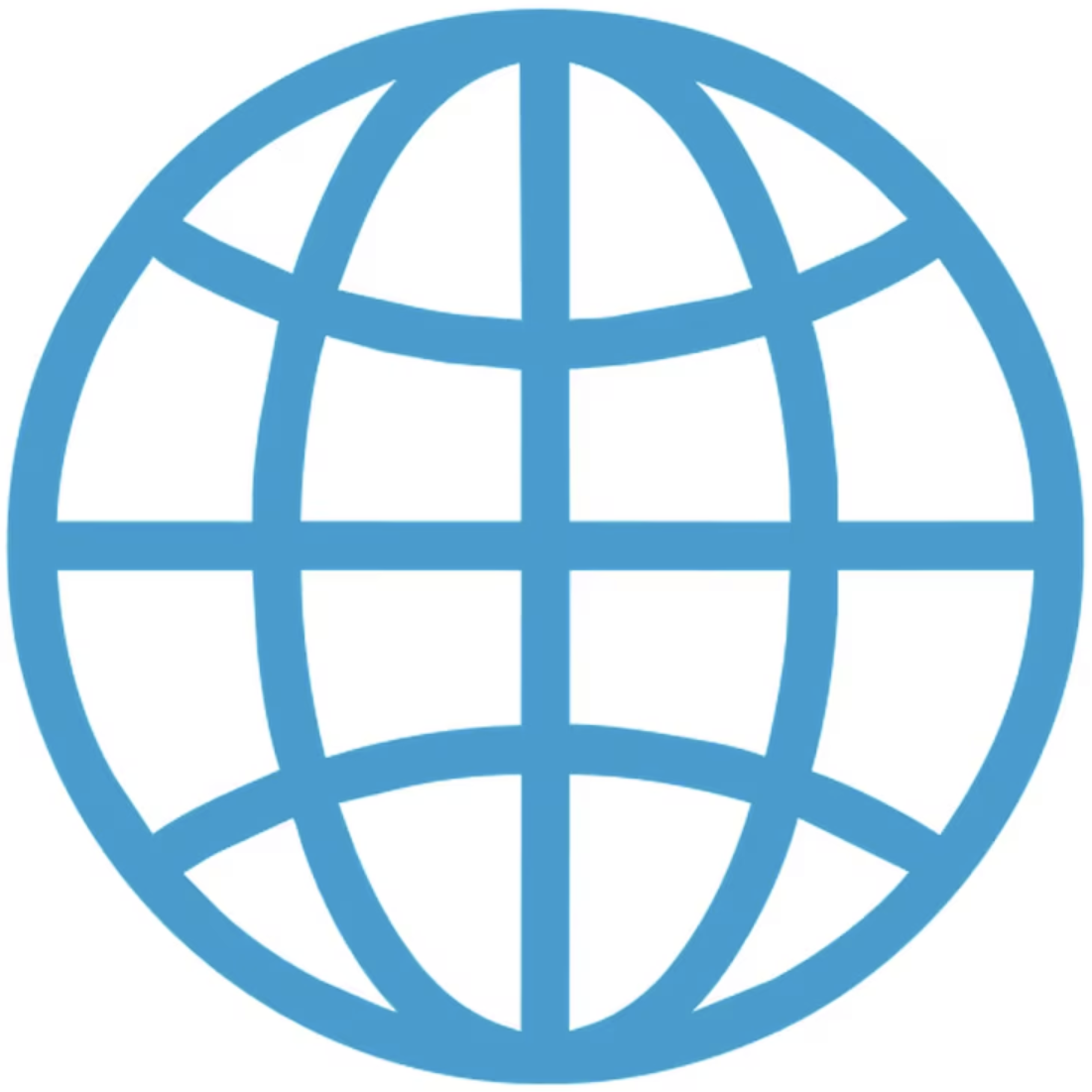}}{\textrm{\textbigcircle}}}

\usetikzlibrary{calc}
\definecolor{barbg}{RGB}{236,239,241}
\newlength\BarW
\newcommand{\barbox}[2]{%
  \begin{tikzpicture}[baseline=0.75ex]
    \pgfmathsetlengthmacro{\w}{#2*\BarW}
    \fill[barbg,rounded corners=0.6pt] (0,0) rectangle (\BarW,1.55ex);
    \fill[#1,rounded corners=0.6pt] (0,0) rectangle (\w,1.55ex);
  \end{tikzpicture}%
}
\newenvironment{BarScale}[2]{%
  \def\ACCmax{#1}\def\GENmax{#2}%
  \newcommand{\accbarv}[1]{\pgfmathsetmacro{\p}{##1/\ACCmax}\barbox{steelBlue}{\p}}
  \newcommand{\genbarv}[1]{\pgfmathsetmacro{\p}{##1/\GENmax}\barbox{oliveGreen}{\p}}
}{}

\definecolor{revision}{RGB}{150, 30, 45}

\usepackage{amsmath,amssymb}
\usepackage[most]{tcolorbox} 

\tcbset{
  examplebox/.style={
    enhanced,
    breakable,
    colback=white,
    colframe=black,
    boxrule=0.6pt,
    arc=3pt,
    outer arc=3pt,
    left=4mm,
    right=4mm,
    top=2mm,
    bottom=2mm,
    colbacktitle=black,
    coltitle=white,
    fonttitle=\bfseries,
  },
  exsmall/.style={
    enhanced,
    breakable,
    colback=gray!10,
    colframe=black,
    boxrule=0.5pt,
    arc=3pt,
    outer arc=3pt,
    left=4mm,
    right=4mm,
    top=2mm,
    bottom=2mm,
  }
}

\title{SwiReasoning: Switch-Thinking in Latent and Explicit for Pareto-Superior Reasoning LLMs}

\author{Dachuan Shi$^1$, Abedelkadir Asi$^2$, Keying Li$^2$, Xiangchi Yuan$^1$, Leyan Pan$^1$, \\ \textbf{Wenke Lee}$^{1\dagger}$, \textbf{Wen Xiao}$^{2\dagger}$ \\
$^1$Georgia Tech \ \ $^2$Microsoft \makebox[0.69\textwidth][r]{\github\ \href{https://github.com/sdc17/SwiReasoning}{github.com/sdc17/SwiReasoning}\,\,\,\, \webiste\ \href{https://swireasoning.github.io/}{swireasoning.github.io}} \\
}

\iclrfinalcopy 
\begin{document}

\maketitle

\vspace{-1em}

\begin{figure}[h]
    \captionsetup{font={small}}
    \centering
    \includegraphics[width=1.0\linewidth]{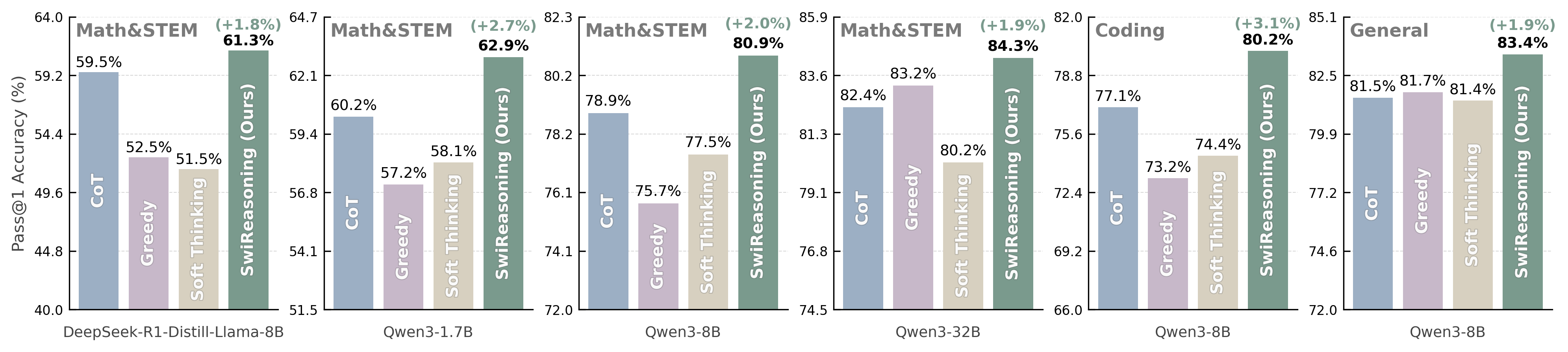}
    \vspace{-1.5em}
    \caption{Pass@1 \textbf{accuracy} under \textit{unlimited} token budgets. Across the Math$\&$STEM, Coding, and General reasoning benchmarks, \swireasoning yields average gains of \textcolor{oliveGreen}{$\boldsymbol{+2.1\%}$}, \textcolor{oliveGreen}{$\boldsymbol{+3.1\%}$}, and \textcolor{oliveGreen}{$\boldsymbol{+1.9\%}$}, respectively.}
    \label{fig:acc}
    
\end{figure}

\vspace{-1em}

\begin{figure}[h]
    \captionsetup{font={small}}
    \centering
    \includegraphics[width=1.0\linewidth]{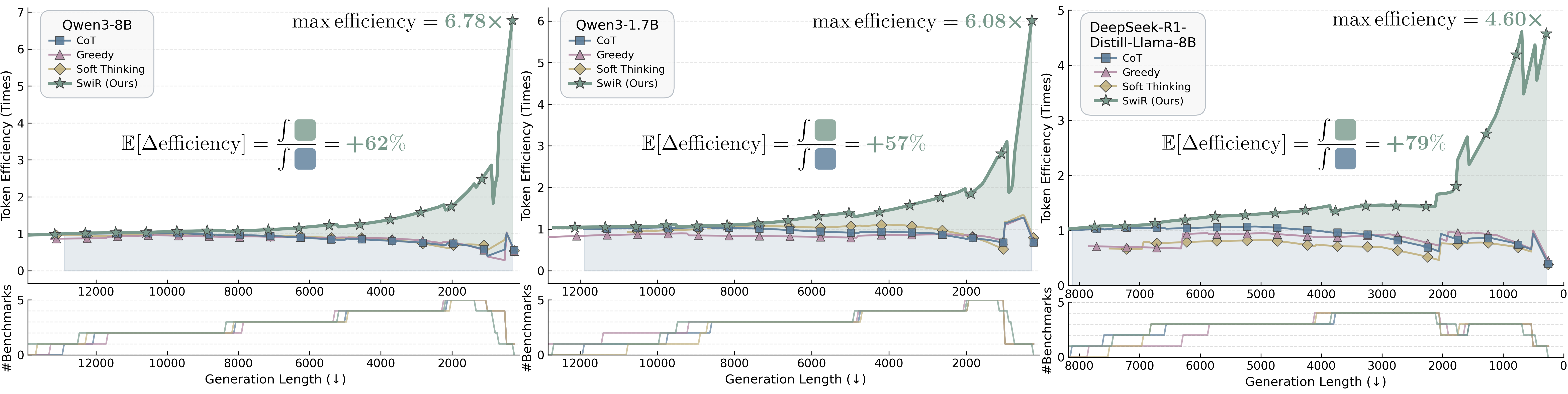}
    \vspace{-1.5em}
    \caption{Token \textbf{efficiency} (accuracy per token compared to standard CoT), under \textit{limited} token budgets. Across reasoning LLM families and sizes, \swireasoning brings average efficiency improvements of up to \textcolor{oliveGreen}{$\boldsymbol{+79\%}$}.}
    \label{fig:eff}
\end{figure}

\begin{abstract}

Recent work shows that, beyond discrete reasoning through explicit chain-of-thought steps, which are limited by the boundaries of natural languages, large language models (LLMs) can also reason continuously in latent space, allowing richer information per step and thereby improving token efficiency.
Despite this promise, latent reasoning still faces two challenges, especially in training-free settings: 1) purely latent reasoning broadens the search distribution by maintaining multiple implicit paths, which diffuses probability mass, introduces noise, and impedes convergence to a single high-confidence solution, thereby hurting accuracy; and 2) overthinking persists even without explicit text, wasting tokens and degrading efficiency.
To address these issues, we introduce \swireasoning, a training-free framework for LLM reasoning which features two key innovations: 
1) \swireasoning dynamically switches between explicit and latent reasoning, guided by block-wise confidence estimated from entropy trends in next-token distributions, to balance exploration and exploitation and promote timely convergence.
2) By limiting the maximum number of thinking-block switches, \swireasoning curbs overthinking and improves token efficiency across varying problem difficulties. 
On widely used mathematics, STEM, coding, and general benchmarks, \swireasoning consistently improves average accuracy by $1.8\%$–$3.1\%$ across reasoning LLMs of different model families and scales. Furthermore, under constrained budgets, \swireasoning improves average token efficiency by $57\%$-$79\%$, with larger gains as budgets tighten.

\end{abstract}

\section{Introduction}

Reasoning is one of the central capabilities of large language models (LLMs)~\citep{yang2025qwen3, qwen2.5, meta2025llama, meta_llama_llama3_3_model_card}. It allows models to tackle complex tasks such as mathematics, science, and programming~\citep{guo2025deepseek, openai2025o3, jaech2024openai, agarwal2025gpt, qwq32b, abdin2025phi, abouelenin2025phi, anthropic2025claude4_system_card, deepmind2023alphazero, deepmind2023imo}, far beyond simple question answering. 

A key limitation of the dominant reasoning approach, explicit chain-of-thought (CoT)~\citep{wolf2020transformers, wei2022chain, yao2023tree, goyalthink, pfau2024let}, lies in the reliance on discrete tokens during inference. In standard CoT decoding, the model commits to a single token at each step, sampled from the predicted distribution. While effective and ensures readability by verbalizing intermediate steps, this discrete process collapses the full probability distribution into a single trajectory, discarding uncertainty and eliminating many potentially useful reasoning paths.

To overcome this bottleneck, recent work has explored an alternative reasoning technique, latent reasoning~\citep{hao2024training, zhang2025softthinking, cheng2024compressed, xu2025softcot, xu2025softcot++, tan2025think}, where the model operates directly in a continuous hidden space instead of a discrete text space. Latent reasoning offers two key advantages over CoT: 1) higher representational bandwidth per step, since hidden vectors can encode richer information than single tokens~\citep{zhu2025survey}; and 2) the ability to preserve multiple reasoning hypotheses implicitly, rather than collapsing them prematurely into one tokenized path~\citep{li2025implicit, chen2025reasoning}.

Latent reasoning can be broadly categorized into training-required and training-free approaches. Training-required ones ~\citep{hao2024training, su2025token, liu2024expediting, shen2025codi, tack2025llm} demand substantial retraining or fine-tuning~\citep{yue2025hybrid, li2025seek, wang2025system, zhu2025reasoning}, making it excessively expensive to apply to large reasoning language models. In contrast, training-free approaches like Soft-Thinking \citep{zhang2025softthinking}, which form a probability-weighted mixture of token embeddings as inputs, operate directly at inference time without incurring additional training costs. Our work focuses on the latter category, which is cost-effective and resource-friendly for deployment in large-scale reasoning models.

Although training-free latent reasoning eliminates the need for costly retraining, operating purely in the latent space also presents significant challenges. First, the model is not explicitly trained to perform long-horizon reasoning with latent inputs. As a result of distributional mismatches, when inference relies solely on latent trajectories, the process is less controlled and can easily drift off course~\citep{chen2025reasoning}. Instead of collapsing into a single path, the model tends to spread probability mass across many implicit reasoning paths. While this preserves multiple hypotheses, it also introduces persistent noise, slows convergence, and ultimately harms reasoning accuracy~\citep{li2025implicit}. Second, the absence of explicit tokens does not necessarily ensure efficiency. In latent space, models may still suffer from repetitive or unnecessarily extended internal deliberations and continuation~\citep{zhang2025softthinking}, essentially overthinking. This prolongs inference and over-consumes tokens, undermining the very efficiency that latent reasoning is meant to improve.  

To address these issues, this paper introduces \swireasoning (abbreviated as \textsc{SwiR}) as a training-free framework for LLM reasoning that alternates between explicit and latent thinking, based on block-wise confidence inferred from entropy trends of next-token distributions, and suppresses overthinking by bounding the number of switches. More specifically, the framework first tracks a reference entropy within each thinking block to reflect block-wise confidence. Rising confidence triggers an explicit switch to consolidate progress along a single path, while sustained uncertainty triggers a latent switch to re-explore in continuous space. Second, a switch count controller caps the number of thinking block transitions and provides early-answer checkpoints, curbing unnecessary latent loops and improving token efficiency across difficulties. 

The proposed framework also benefits from reintroducing diversity by sampling in an explicit thinking block when compared to pure latent thinking. Even though motivated differently, \swireasoning resonates with the concurrent observation of \citet{liang2025singlethread} that introducing stochasticity benefits latent reasoning, but we achieve this via a distinct mode switch mechanism rather than injecting distributions with randomness.

Our contributions are summarized as follows:
\begin{itemize}[leftmargin=*,labelsep=0.5em]
  \item We propose \swireasoning, a training-free reasoning framework that dynamically alternates between explicit and latent thinking based on confidence signals, thereby exploiting the expressivity of latent thinking without sacrificing the stability of explicit thinking. 
  \item We introduce a switch count control mechanism that caps the number of transitions, enabling early answering based on partial reasoning trajectories at switch boundaries. This effectively suppresses overthinking and improves token efficiency under limited budgets.
  \item We extensively validate the effectiveness of \swireasoning on mathematics, STEM, coding, and general reasoning domains across multiple benchmarks, model families, and sizes, demonstrating consistent gains in both accuracy and token efficiency over training-free baselines.
\end{itemize}

\section{Related Work}

\paragraph{Explicit LLM Reasoning.}

Reasoning via explicit intermediate text has been extensively studied. Chain-of-thought (CoT) prompting elicits stepwise rationales that improve reasoning accuracy by decomposing problems into natural-language sub-steps~\citep{kojima2022large, wei2022chain}. Subsequent work increases robustness by aggregating multiple CoT trajectories through self-consistency ~\citep{wang2022self}. Search- and tool-augmented variants further expand the exploration space, such as Tree-of-thought that branches over partial rationales~\citep{yao2023tree} and interleaving reasoning and actions with external tools and environments~\citep{yao2023react}. Least-to-most prompting progressively solves subproblems to reduce reasoning load and mitigate error accumulation ~\citep{zhou2022least}. These approaches operate purely in the discrete token space and therefore commit to a single token at each step. While readable, the discretizations in explicit reasoning discard alternative hypotheses early, and restrict the information bandwidth per step \citep{zhu2025survey}.

\paragraph{Latent LLM Reasoning.}

Latent reasoning operates in the continuous representation space rather than discrete natural language space used by explicit reasoning. Prior work can be broadly grouped into two categories. 1) Training-required approaches modify pretraining~\citep{tack2025llm, zeng2025pretraining} or fine-tuning objectives~\citep{tan2025think, wang2025system, wang2025, jiang2025, wu2025parallel, yue2025hybrid, li2025seek, shen2025codi, xu2025softcot} to supervise hidden-state trajectories or to endow models with latent-planning skills. 2) Training-free approaches~\citep{zhang2025softthinking, liang2025singlethread} intervene only at inference time by manipulating hidden representations or probability distributions without updating model weights. Our work belongs to the training-free category but differs from prior single-mode methods. Instead of remaining purely latent, \swireasoning dynamically switches between latent and explicit reasoning based on entropy-trend confidence, and further regulates the number of switches through a count controller to suppress overthinking and improve efficiency.

\section{Methodology}

\subsection{SwiReasoning Overview}

\begin{figure}[t]
    \captionsetup{font={small}}
    \centering
    \includegraphics[width=1.0\linewidth]{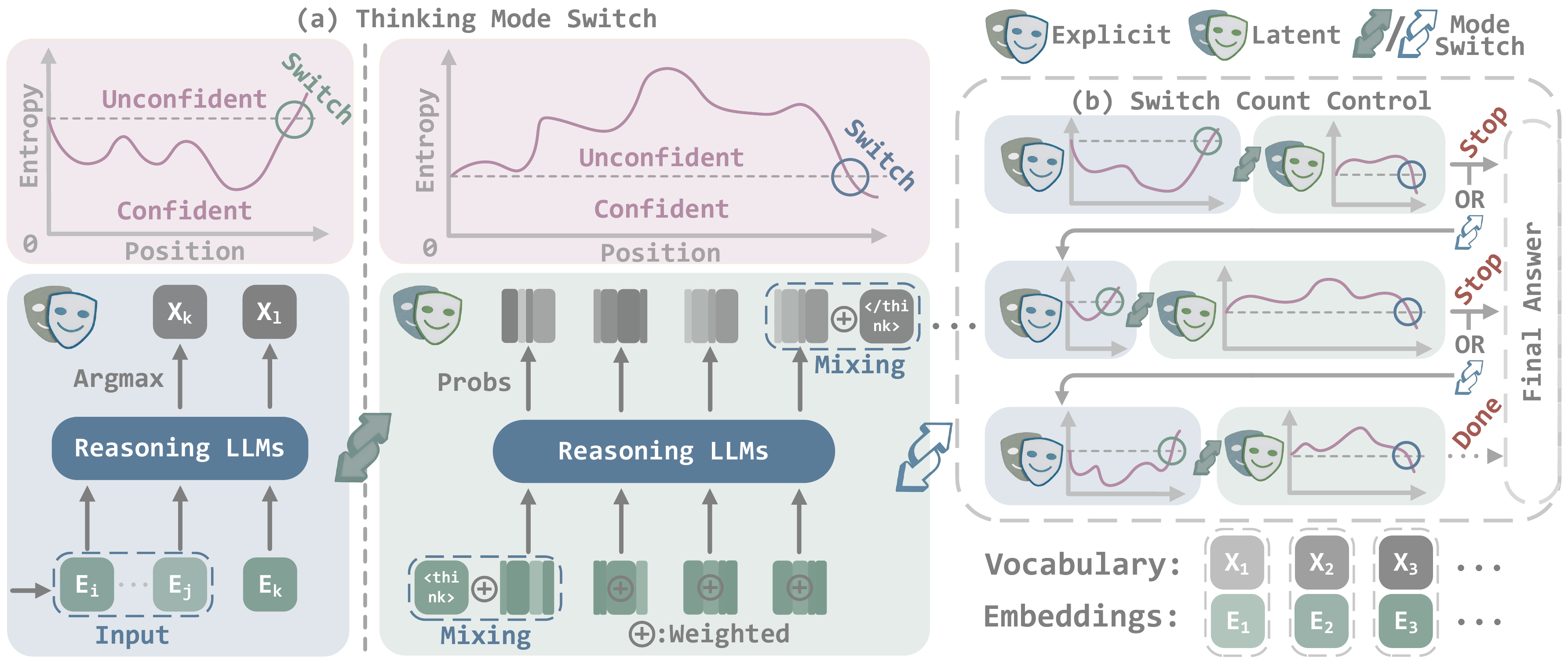}
    \vspace{-1em}
    \caption{\swireasoning framework. (a) Dynamic mode switching alternates between explicit and latent thinking based on block-wise confidence estimated from entropy trends. (b) A switch count control mechanism limits the maximum number of thinking-block transitions, suppressing overthinking before the final answer. }
    \label{fig:method}
\end{figure}

As shown in Fig.~\ref{fig:method}, \swireasoning is a training-free framework that dynamically alternates between explicit and latent reasoning. The number of switches is regulated to suppress overthinking and improve token efficiency. 
Sec.~\ref{sec:prelim} presents the preliminaries of explicit and latent reasoning, Sec.~\ref{sec:dynamic_switch} details the design of the dynamic switch, and Sec.~\ref{sec:switch_control} discusses the switch count control mechanism. Implementation details are provided in Appendix \ref{supp_imp}.

\subsection{Preliminary: Explicit and Training-Free Latent Thinking} 
\label{sec:prelim}

\paragraph{Explicit Thinking.}
Let $V$ be a vocabulary and $p_\theta(x_t \mid x_{<t})$ an LLM over $V$ with parameters $\theta$.
Given a question $q$, the model produces a reasoning trace $r_{1:T}\in V^T$ followed by a final answer $a_{1:U}\in V^U$.
We write the concatenated sequence as
\[
x_{1:(\lvert q\rvert + T + U)} \;=\; \big[q,\, r_{1:T},\, a_{1:U}\big], 
\]
At inference, decoding proceeds by repeatedly choosing a token $x_t$ from the predictive distribution $p_\theta(\cdot \mid x_{<t})$ according to a policy $\pi_t(\cdot)$, \textit{e.g.},
\[
x_t \sim \pi_t(\cdot) 
\;\;\text{with}\;\;
\pi_t = 
\begin{cases}
\text{Greedy: } \arg\max_{v\in V} p_\theta(v\mid x_{<t}),\\[2pt]
\text{Sampling: } \text{Top-}k/\text{Top-}p \text{ with temperature }\tau.
\end{cases}
\]
The reasoning phase stops when a termination condition is met, \textit{e.g.}, generating $\langle\text{/think}\rangle$, after which the answer tokens $a_{1:U}$ are decoded in the same manner.
While explicit reasoning improves reliability by externalizing intermediate steps, its hard policy $\pi_t(\cdot)$ collapses the full distribution to a single discrete decision at each step, \textit{i.e.}, discards information in $p_\theta(\cdot\mid x_{<t})$ beyond the chosen token.

\paragraph{Training-Free Latent Thinking.}
It replaces the hard policy $\pi_t(\cdot)$ by a continuous surrogate that preserves distributional information.
Let $E\in\mathbb{R}^{\lvert V\rvert\times d}$ denote the token embedding matrix with rows $e^{(v)}\!\in\!\mathbb{R}^d$.
At step $t$, the model yields logits $\ell_t\in\mathbb{R}^{|V|}$ and $p_t=\mathrm{softmax}(\ell_t)$.
Given the next-token distribution $p_t := p_\theta(\cdot \mid x_{<t})\in\Delta^{\lvert V\rvert-1}$, it forms a soft embedding
\begin{equation}
\label{eq:soft-embed}
\tilde{e}_t 
\;=\; \sum_{v\in V} p_t[v]\, e^{(v)} 
\;\in\; \mathbb{R}^d,
\end{equation}
and feeds $\tilde{e}_t$ back to the model as the next input representation, rather than committing to an explicit token by $\pi_t(\cdot)$. Upon the thinking phase being complete, the policy reverts to $\pi_t(\cdot)$ for answer generation.
The convexity of Eq.~\ref{eq:soft-embed} ensures $\tilde{e}_t$ lies in the embedding hull of $E$, retaining all first-order uncertainty in $p_t$, which reduces information discards and increases robustness to local noise.

\subsection{Dynamic Switch between Explicit and Latent Thinking}
\label{sec:dynamic_switch}

Remaining in a single mode throughout reasoning is inherently suboptimal: explicit thinking provides readability but may discard useful information beyond chosen tokens, while latent thinking preserves richer signals but can drift into noise and reduce accuracy. Our key insight is that \textit{reasoning should switch modes based on confidence}. Latent reasoning enables exploration across multiple potential continuations when confidence is low, and explicit reasoning encourages convergence when confidence is high, striking a balance that supports broad exploration while maintaining accuracy.

\paragraph{Mode Switch Criterion.} 
We refer to the reasoning content between two consecutive switches as a thinking block and estimate its confidence by entropy
\(
H_t \!=\! -\sum_{v} p_t[v]\log p_t[v].
\)
Let $\bar H$ denote the reference entropy of the current block, which is initialized at the first step of the block and refreshed when a mode switch happens. We use a criterion that converts local entropy trends into decisions:
\begin{align}
\label{eq:s2n}
\text{Latent}\!\to\!\text{Explicit}:\quad &
\left(H_t < \bar H\right) \;\;\text{(confidence rises)} ,
\\
\label{eq:n2s}
\text{Explicit}\!\to\!\text{Latent}:\quad &
\left(H_t > \bar H\right) \;\;\text{(confidence drops)},
\end{align}

\paragraph{Switch Window Size.}
\label{sec:window}
To avoid oscillations, we impose dwell windows upon the mode switch criterion.
Formally, with mode variable $m_t\in\{\text{Explicit},\text{Latent}\}$ and dwell step counter $\Delta t$, we have 
\[
m_{t+1}=
\begin{cases}
\text{Explicit}, & m_t=\text{Latent} \wedge (H_t<\bar H) \wedge (\Delta t\ge W_{\text{L}\to\text{E}}), \\
\text{Latent}, & m_t=\text{Explicit} \wedge (H_t>\bar H) \wedge (\Delta t\ge W_{\text{E}\to\text{L}}), \\
m_t, & \text{otherwise}.
\end{cases}
\]
We reset $\bar H\!\leftarrow\!H_t,\,\Delta t\!\leftarrow\!0$ upon any switch, \textit{i.e.}, $m_{t+1}\ne m_t$. Otherwise, we update $\Delta t\!\leftarrow\!\Delta t{+}1$. In practice, $W_{\text{L}\to\text{E}} = 0$ while $W_{\text{E}\to\text{L}}$ is positive, \textit{i.e.}, a Latent$\to$Explicit switch may occur immediately when $H_t$ dips, whereas an Explicit$\to$Latent switch requires staying for at least $W_{\text{E}\to\text{L}}$ steps. 

The key intuition behind the asymmetric design is that \textit{two modes play different roles in reasoning}. 
Latent reasoning is inherently divergent, allowing for rich exploration. However, prolonging the latent phase after confidence has recovered is counterproductive. It increases the risks of introducing spurious signals that may mislead the model. Therefore, once confidence rises, an immediate switch back to explicit reasoning is necessary to consolidate progress onto a single coherent trajectory. 

In contrast, explicit reasoning is convergent, gradually unfolding a chain-of-thought where each token incrementally extends the current logical path. 
If the model were allowed to switch back to latent reasoning at the first sign of an entropy fluctuation, spurious short-term uncertainty could trigger oscillations. The dwell window $W_{E \rightarrow L}$ ensures that explicit reasoning is given sufficient opportunity to stabilize and accumulate meaningful structure. 

\paragraph{Thinking-Related Signal Mixing.}
\label{sec:mixing}
To better align mode switches with the LLMs' learned reasoning patterns, we blend the embeddings of thinking-related signal tokens, \textit{e.g.}, \texttt{\textless think\textgreater} and \texttt{\textless /think\textgreater}, when a switch occurs.
Let $e_{\langle \text{think}\rangle}$ and $e_{\langle\!/\text{think}\rangle}$ denote their embeddings.
At the entrance to a latent thinking block, we bias the first latent step $t^\star$ toward ``begin thinking'' by
\begin{equation}
\label{eq:enter-latent}
\tilde e_{t^\star}\;\leftarrow\; \alpha_{t^\star}\!\cdot\!\tilde e_{t^\star}
\;+\; (1-\alpha_{t^\star})\!\cdot\! e_{\langle \text{think}\rangle},
\qquad \alpha_{t^\star}\in[0,1],
\end{equation}
and at the exit to an explicit thinking block, we bias the first explicit step $t^\dagger$ toward ``end thinking'' 
\begin{equation}
\label{eq:enter-explicit}
\tilde e_{t^\dagger}\;\leftarrow\;
\beta_{t^\dagger}\!\cdot\!\tilde e_{{t^\dagger}}
\;+\; (1-\beta_{t^\dagger})\!\cdot\! e_{\langle\!/\text{think}\rangle},
\qquad \beta_{t^\dagger}\in[0,1],
\end{equation}
which encourages the model to close the latent phase and move on to answer production.
In practice, we schedule $\alpha_t=\alpha_0+(1-\alpha_0)\frac{t}{T_{\max}}$ and
$\beta_t=\beta_0+(1-\beta_0)\frac{t}{T_{\max}}$, where $T_{\max}$ is a predefined maximum generation length,
and apply Eq.~\ref{eq:enter-latent} or Eq.~\ref{eq:enter-explicit} only at the steps when switches occur.

\subsection{Overthinking Suppression by Switch Count Control}
\label{sec:switch_control}

Even with confidence-aware switching, reasoning LLMs may still overthink. Therefore, we place a bound on the total number of Latent$\to$Explicit switches. Our key insight is that each switch naturally marks \textit{the end of a thinking block where partial reasoning trajectories have been consolidated}, which may already contain sufficient evidence for arriving at a reasonable solution. Under limited budgets, generating answers at these natural checkpoints can make use of partial reasoning trajectories, offering a chance to obtain correct predictions earlier without consuming additional tokens.

\paragraph{Counter and Triggers.}
Let $C_t$ count completed $\text{Latent}\!\to\!\text{Explicit}$ switches up to step $t$.
Given a user-specified budget $C_{\max}$, we define two triggers:

\begin{itemize}[leftmargin=12pt]
\item \textbf{Convergence trigger} (at $\frac{1}{2}C_{\max} \leq C_t \leq C_{\max}$ on $\text{Latent}\!\to\!\text{Explicit}$ transitions):
force the next token to be $\langle\!/\text{think}\rangle$. The convergence trigger is to encourage rather than enforce the end of the thinking process and the start of converging to an answer based on partial reasoning trajectories.

\item \textbf{Termination trigger} (at $C_t > C_{\max}$ on a subsequent $\text{Latent}\!\to\!\text{Explicit}$ transition):
inject a concise answer prefix $s_{\text{final}}$, ``$\langle\!/\text{think}\rangle$\text{\textbackslash n\textbackslash n The final answer is}'', then allow at most $B$ additional tokens for the final answer. The termination trigger is to enforce an immediate answer generation.

\end{itemize}

Triggers are implemented as short injection queues that overwrite future-generated tokens. 
Formally, let $\mathcal{Q}_t$ be the per-sample injection queue.
When a convergence or termination trigger fires, we set $\mathcal{Q}_t\!\gets\![\text{ID}(\langle\!/\text{think}\rangle)]$ or $[\text{ID}(s_{\text{final}})]$.
At the next step, if $\mathcal{Q}_t\neq\emptyset$, we deterministically set $x_t\!\leftarrow\!\mathcal{Q}_t.\mathrm{pop}()$.
For the termination trigger, we also start a budget counter $b_t\!=\!B$ and decrement it each step after the termination trigger fires. Decoding will be terminated once $b_t\!=\!0$.

\section{Experiments}

\subsection{Experimental Settings}

\paragraph{Models.} 
We evaluate \swireasoning on four recent reasoning LLMs: DeepSeek-R1-Distill-Llama-8B~\citep{guo2025deepseek}, Qwen3-1.7B~\citep{yang2025qwen3}, Qwen3-8B~\citep{yang2025qwen3}, and Qwen3-32B~\citep{yang2025qwen3}. 
This selection helps us validate the effectiveness of \swireasoning across different model families, model scales, and training paradigms.

\paragraph{Domains and Benchmarks.} We evaluate \swireasoning on 11 benchmarks spanning four domains: mathematical reasoning (GSM8K~\citep{cobbe2021training}, Math500~\citep{hendrycks2021measuring}, AIME 2024~\citep{hf_aime24_dataset}, AIME 2025~\citep{hf_aime25_dataset}); STEM reasoning (GPQA Diamond~\citep{rein2024gpqa}); coding reasoning (HumanEval~\citep{chen2021evaluating}, LeetCode-Contest~\citep{guo2024deepseekcoder}, MBPP~\citep{austin2021program}, LiveCodeBench~\citep{jain2024livecodebench}); and general reasoning (2WikiMultihopQA~\citep{ho2020constructing}, CommonsenseQA~\citep{talmor2019commonsenseqa}).


\paragraph{Baselines.} We compare \swireasoning that dynamically switches between thinking modes against three baselines with a single thinking mode, including 1) explicit thinking alone: standard CoT reasoning with sampling, standard CoT reasoning with greedy decoding, and 2) training-free latent thinking alone: Soft Thinking ~\citep{zhang2025softthinking}. 

\paragraph{Metrics.}
We use the Pass@1 metric to evaluate reasoning accuracy, and token efficiency $E$ to assess the level of reasoning efficiency. Let $\mathrm{Acc}_m(\ell)\in[0,1]$ denote the accuracy of method $m$ when using $\ell$ generated tokens. Its plain token efficiency is the accuracy gained per token,
\[
PE_m(\ell) \;=\; \frac{\mathrm{Acc}_m(\ell)}{\ell}.
\]
To express efficiency in units relative to the standard CoT, we normalize it by the CoT's plain token efficiency when the highest accuracy is achieved. Specifically, if CoT achieves its highest accuracy $\mathrm{Acc}_{\mathrm{CoT}}^{\star}$ using $\ell_{\mathrm{CoT}}^{\star}$ tokens, denote
\(
PE_{\mathrm{CoT}}^{\star}
=
\frac{\mathrm{Acc}_{\mathrm{CoT}}^{\star}}{\ell_{\mathrm{CoT}}^{\star}}.
\)
The token efficiency of $m$ is defined as
\[
\mathrm{E}_m(\ell)
\;=\;
\frac{PE_m(\ell)}{PE_{\mathrm{CoT}}^{\star}}
\;=\;
\frac{\mathrm{Acc}_m(\ell)/\ell}{\mathrm{Acc}_{\mathrm{CoT}}^{\star}/\ell_{\mathrm{CoT}}^{\star}}.
\]

And the average efficiency gain of method $m$ over CoT is
\[
\mathbb{E}[\Delta \mathrm{E}_{\text{m}}]
\;=\;
\frac{\int (\mathrm{E}_{\text{m}}(\ell) - \mathrm{E}_{\text{CoT}}(\ell))\, d\ell}
     {\int \mathrm{E}_{\text{CoT}}(\ell)\, d\ell}.
\]

\begin{table*}[t]
\small
\centering
\captionsetup{font={small}}
\setlength{\tabcolsep}{0.42mm}
\caption{Comparison of \swireasoning and CoT with sampling, CoT with greedy decoding, and Soft Thinking on mathematics and STEM benchmarks. \swireasoning improves accuracy by \textcolor{oliveGreen}{$\boldsymbol{+2.17\%}$} on average.}
\label{tab:accuracy_unlimited}

\begin{tabular}{l r l  r l  r l  r l  r l  r l}
\toprule
\multirow{3}{*}[2pt]{\makecell{\textbf{Method}}} &
\multicolumn{2}{c}{\makecell{GSM8K}} &
\multicolumn{2}{c}{\makecell{MATH\\500}} &
\multicolumn{2}{c}{\makecell{GPQA\\Diamond}} &
\multicolumn{2}{c}{\makecell{AIME\\2024}} &
\multicolumn{2}{c}{\makecell{AIME\\2025}} &
\multicolumn{2}{c}{\makecell{\textbf{Average}}} \\
\cmidrule{2-11}\cmidrule(l){12-13}

\multicolumn{13}{c}{\textit{Qwen3-8B}~\citep{yang2025qwen3}} \\
\midrule
CoT               & 95.60 & 
                  & 96.00 & 
                  & 59.60 & 
                  & 75.83 & 
                  & 67.50 & 
                  & 78.91 & \multicolumn{1}{c}{\scriptsize\textcolor{gray}{$+0.00$}} \\
CoT (Greedy)      & 95.68 & 
                  & 96.40 & 
                  & 56.57 & 
                  & 70.00 & 
                  & 60.00 & 
                  & 75.73 & \multicolumn{1}{c}{\scriptsize\textcolor{mauve}{$-3.18$}} \\
Soft Thinking     & 95.38 & 
                  & 96.00 & 
                  & 59.60 & 
                  & 67.92 & 
                  & 68.33 & 
                  & 77.45 & \multicolumn{1}{c}{\scriptsize\textcolor{mauve}{$-1.46$}} \\
\rowcolor{gray!20}\textit{\textbf{SwiR} (Ours)}       & \textbf{96.06} & \Deltabar{ 0.46}
                  & \textbf{98.40} & \Deltabar{ 2.40}
                  & \textbf{61.11} & \Deltabar{ 1.51}
                  & \textbf{79.17} & \Deltabar{ 3.34}
                  & \textbf{70.00} & \Deltabar{ 2.50}
                  & \textbf{80.94} & \Deltabar{ 2.03} \\
\midrule

\multicolumn{13}{c}{\textit{Qwen3-1.7B}~\citep{yang2025qwen3}} \\
\midrule
CoT               & 90.44 & 
                  & 92.00 & 
                  & 39.39 & 
                  & 45.83 & 
                  & 33.33 & 
                  & 60.20 & \multicolumn{1}{c}{\scriptsize\textcolor{gray}{$+0.00$}} \\
CoT (Greedy)      & 89.61 & 
                  & 91.00 & 
                  & 31.82 & 
                  & 40.00 & 
                  & 33.33 & 
                  & 57.15 & \multicolumn{1}{c}{\scriptsize\textcolor{mauve}{$-3.05$}} \\
Soft Thinking     & 90.30 & 
                  & 90.60 & 
                  & 34.34 & 
                  & 38.75 & 
                  & 36.67 & 
                  & 58.13 & \multicolumn{1}{c}{\scriptsize\textcolor{mauve}{$-2.07$}} \\
\rowcolor{gray!20}\textit{\textbf{SwiR} (Ours)}       & \textbf{90.83} & \Deltabar{ 0.39}
                  & \textbf{93.00} & \Deltabar{ 1.00}
                  & \textbf{41.41} & \Deltabar{ 2.02}
                  & \textbf{50.83} & \Deltabar{ 5.00}
                  & \textbf{38.33} & \Deltabar{ 5.00}
                  & \textbf{62.88} & \Deltabar{ 2.68} \\
\midrule

\multicolumn{13}{c}{\textit{DeepSeek-R1-Distill-Llama-8B}~\citep{guo2025deepseek}} \\
\midrule
CoT               & 89.46 & 
                  & 91.40 & 
                  & 46.46 & 
                  & 43.75 & 
                  & 26.25 & 
                  & 59.46 & \multicolumn{1}{c}{\scriptsize\textcolor{gray}{$+0.00$}} \\
CoT (Greedy)      & 85.82 & 
                  & 84.80 & 
                  & 31.81 & 
                  & 30.00 & 
                  & 30.00 & 
                  & 52.49 & \multicolumn{1}{c}{\scriptsize\textcolor{mauve}{$-6.97$}} \\
Soft Thinking     & 85.90 & 
                  & 83.80 & 
                  & 33.33 & 
                  & 34.17 & 
                  & 20.42 & 
                  & 51.52 & \multicolumn{1}{c}{\scriptsize\textcolor{mauve}{$-7.94$}} \\
\rowcolor{gray!20}\textit{\textbf{SwiR} (Ours)}       & \textbf{90.07} & \Deltabar{ 0.61}
                  & \textbf{92.00} & \Deltabar{ 0.60}
                  & \textbf{47.98} & \Deltabar{ 1.52}
                  & \textbf{45.00} & \Deltabar{ 1.25}
                  & \textbf{31.25} & \Deltabar{ 5.00}
                  & \textbf{61.26} & \Deltabar{ 1.80} \\
\bottomrule
\end{tabular}
\end{table*}

\subsection{Reasoning Accuracy under unlimited Token Budgets}

We first evaluate \swireasoning in the setting where token budgets are set large enough to ensure that each method is allowed to conduct sufficient thinking (refer to Appendix~\ref{supp_benchmark} for detailed settings). 
Fig.~\ref{fig:acc} and Tab.~\ref{tab:accuracy_unlimited} report the highest attainable accuracies across mathematics (GSM8K, MATH500, AIME24, AIME25) and STEM (GPQA Diamond) benchmarks under this setting. 
Across different model families of varying sizes, \swireasoning consistently achieves higher Pass@1 accuracy than CoT with sampling, CoT with greedy decoding, and Soft Thinking. 

Our observation is that improvements are most pronounced on the more challenging benchmarks. 
For instance, on AIME24/AIME25, which require deep deductive reasoning and are widely regarded as more difficult, our method yields absolute gains of $3.34\%/2.50\%$ on Qwen3-8B, and $5.00\%/5.00\%$ on Qwen3-1.7B. 
These margins substantially exceed those observed on GSM8K or MATH500 with lower difficulty, suggesting that the proposed switching mechanism is particularly beneficial when problems involve long reasoning chains or higher uncertainty from the LLM's perspective. Overall, the accuracy results under unlimited token budgets highlight the strength of \swireasoning in better addressing reasoning tasks compared to single-mode approaches.

\begin{figure}[t]
    \captionsetup{font={small}}
    \centering
    \includegraphics[width=1.0\linewidth]{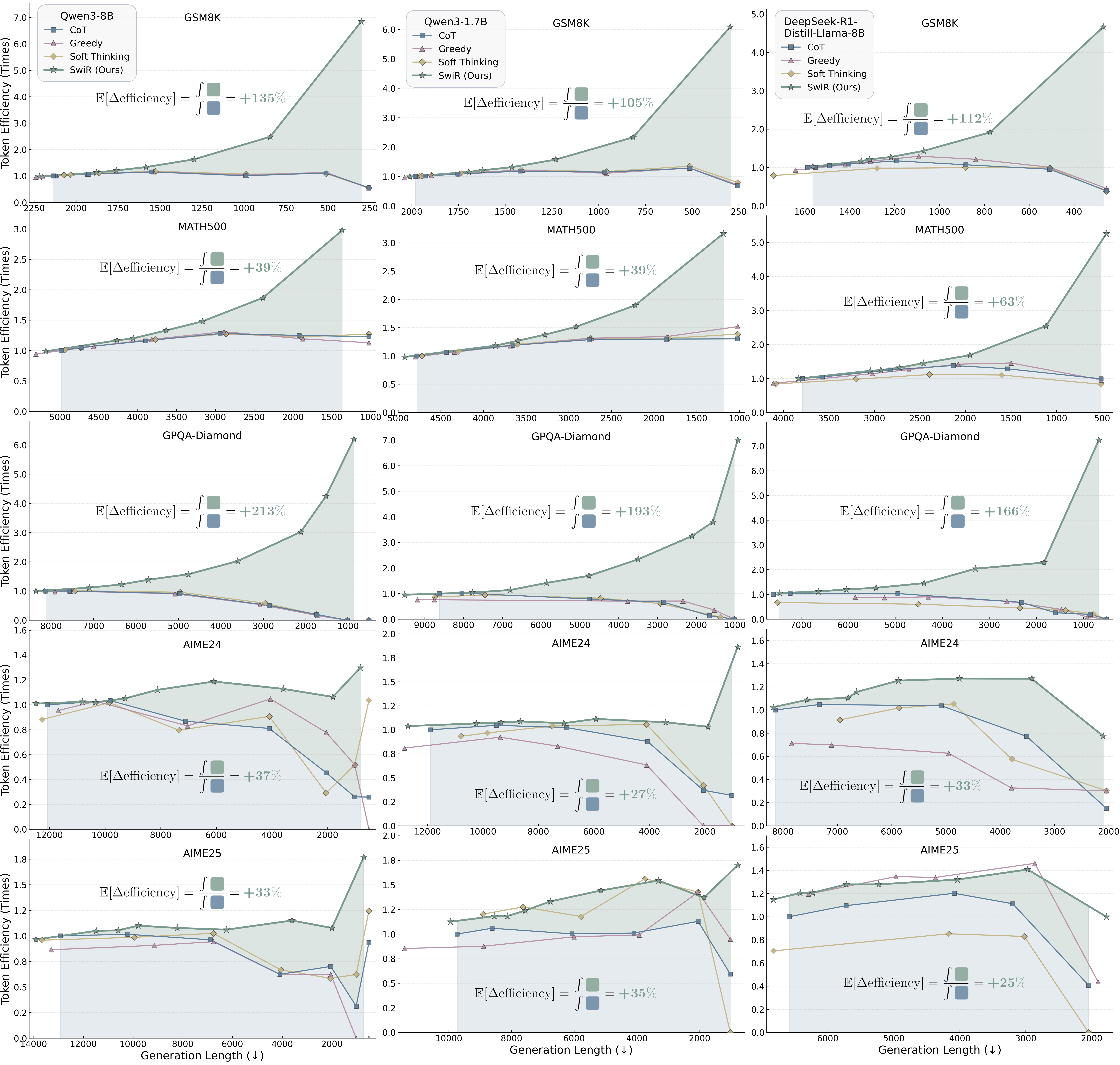}
    \vspace{-1.6em}
    \caption{Token efficiency comparisons. \swireasoning achieves the highest token efficiency throughout all token budgets in 13 out of 15 evaluations, with an efficiency improvement of \textcolor{oliveGreen}{$\boldsymbol{+84\%}$} over CoT on average.}
    \label{fig:eff-single}
\end{figure}

\subsection{Token Efficiency under limited Token Budgets}
\label{sec:exp_efficiency}

Across models and benchmarks, \swireasoning consistently attains improved Pareto frontiers. As shown in Fig.~\ref{fig:eff}, the peak efficiency gains range between $4.6\times$ and $6.8\times$ over CoT depending on the model size. These improvements are not confined to a single budget: the area-under-curve (AUC) advantage persists across a broad range of small to moderate budgets.

One observation from Fig.~\ref{fig:eff-single} is that the relatively large average efficiency gains occur on GSM8K, MATH500, and GPQA Diamond across three models (up to $213\%$ AUC improvements in the per-benchmark panels). These tasks contain many instances with lower difficulty, which benefit most from our overthinking suppression design to obtain the correct answer with partial reasoning trajectories. In contrast, on AIME24/25, the efficiency gaps are smaller, while the accuracy gains with unlimited budgets are larger. This asymmetry is expected: the harder the problem is, the more difficult it is to predict a correct answer with unfinished reasoning trajectories. Overall, token efficiency results under limited budgets substantiate the advantage of \swireasoning in gaining accuracy more efficiently as budgets tighten compared to baseline methods.

\begin{figure*}[ht]
    \vspace{-0.5em}
    \captionsetup{font={small}}
    \centering
    \includegraphics[width=0.9\linewidth]{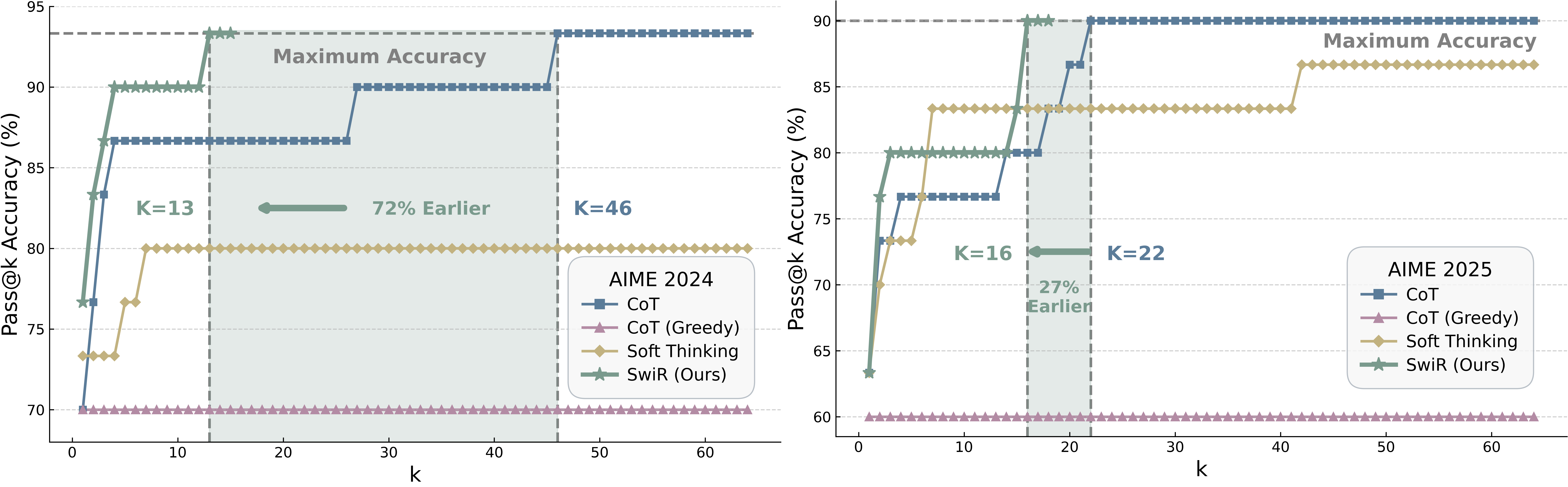}
    \vspace{-1.1em}
    \caption{Pass@k accuracy ($k \in [1,64]$) evaluation with Qwen3-8B on AIME 2024 and 2025 benchmarks. \swireasoning achieves maximum reasoning accuracies \textcolor{oliveGreen}{$\boldsymbol{+50\%}$} earlier compared to CoT on average.}
    \label{fig:passk}
    \vspace{-0.9em}
\end{figure*}

\subsection{Evaluation with Pass@k Accuracy}

In addition to Pass@1 accuracy, we also measure Pass@k accuracy, where $k\in[1,64]$ on Qwen3-8B. Fig.~\ref{fig:passk} shows that \swireasoning reaches its maximal accuracy with significantly smaller $k$ than baselines. Define $k^\star$ as the smallest $k$ achieving the method's peak. On AIME24, we observe $k^\star=13$ for \swireasoning versus $46$ for CoT (about $72\%$ fewer samples), and on AIME25, $k^\star=16$ versus $22$ (about $27\%$ fewer samples). In addition to the faster growth of the curve than CoT, \swireasoning also exhibits 1) a steeper initial slope at small $k$ (higher "per-sample yield"), and 2) a higher eventual ceiling than Soft Thinking and greedy CoT, indicating better correctness and diversity simultaneously. Overall, Pass@k accuracy results indicate that \swireasoning is particularly attractive for budgeted evaluation settings where $k$ cannot be large.

\subsection{Ablation Studies}

\paragraph{Switch Window Size.} 

\begin{table}[t]
\centering
\scriptsize
\captionsetup{font={small}}
\caption{Ablations on $\alpha_0$ and $\beta_0$ for signal mixing. \textcolor{oliveGreen}{\textbf{Greener}} indicates better performance within each column.}
\label{tab:ablation-alpha-beta}
\setlength{\tabcolsep}{2.8pt}
\begin{tabular}{lcccccclcccccc}
\toprule
$\mathbf{\alpha_0}$ & \makecell{GSM8K} & \makecell{MATH \\ 500} & \makecell{GPQA \\ Diamond} & \makecell{AIME \\ 2024} & \makecell{AIME \\ 2025} & \makecell{\textbf{Average}} &
$\mathbf{\beta_0}$ & \makecell{GSM8K} & \makecell{MATH \\ 500} & \makecell{GPQA \\ Diamond} & \makecell{AIME \\ 2024} & \makecell{AIME \\ 2025} & \makecell{\textbf{Average}} \\
\cmidrule(r){1-1} \cmidrule(r){2-6} \cmidrule(r){7-7} \cmidrule{8-8} \cmidrule(lr){9-13} \cmidrule{14-14}
0.0  & \cellcolor{oliveGreen!20!white}89.23\%
     & \cellcolor{oliveGreen!20!white}89.80\%
     & \cellcolor{oliveGreen!35!white}35.86\%
     & \cellcolor{oliveGreen!43!white}46.67\%
     & \cellcolor{oliveGreen!50!white}35.00\%
     & \cellcolor{oliveGreen!20!white}59.31\%
  & 0.0 & \cellcolor{oliveGreen!20!white}81.50\%
     & \cellcolor{oliveGreen!20!white}67.20\%
     & \cellcolor{oliveGreen!23!white}28.79\%
     & \cellcolor{oliveGreen!20!white}8.33\%
     & \cellcolor{oliveGreen!21!white}9.17\%
     & \cellcolor{oliveGreen!20!white}39.00\% \\

0.1  & \cellcolor{oliveGreen!47!white}89.84\%
     & \cellcolor{oliveGreen!46!white}91.00\%
     & \cellcolor{oliveGreen!40!white}36.36\%
     & \cellcolor{oliveGreen!59!white}46.25\%
     & \cellcolor{oliveGreen!86!white}36.25\%
     & \cellcolor{oliveGreen!68!white}59.94\%
  & 0.1 & \cellcolor{oliveGreen!23!white}81.88\%
     & \cellcolor{oliveGreen!28!white}70.20\%
     & \cellcolor{oliveGreen!38!white}31.82\%
     & \cellcolor{oliveGreen!26!white}11.67\%
     & \cellcolor{oliveGreen!20!white}8.75\%
     & \cellcolor{oliveGreen!25!white}40.86\% \\

0.2  & \cellcolor{oliveGreen!70!white}90.37\%
     & \cellcolor{oliveGreen!59!white}91.60\%
     & \cellcolor{oliveGreen!25!white}34.85\%
     & \cellcolor{oliveGreen!60!white}46.25\%
     & \cellcolor{oliveGreen!85!white}37.50\%
     & \cellcolor{oliveGreen!70!white}60.11\%
  & 0.2 & \cellcolor{oliveGreen!25!white}82.11\%
     & \cellcolor{oliveGreen!29!white}70.60\%
     & \cellcolor{oliveGreen!20!white}28.28\%
     & \cellcolor{oliveGreen!30!white}14.17\%
     & \cellcolor{oliveGreen!21!white}9.17\%
     & \cellcolor{oliveGreen!25!white}40.87\% \\

0.3  & \cellcolor{oliveGreen!73!white}90.45\%
     & \cellcolor{oliveGreen!59!white}91.60\%
     & \cellcolor{oliveGreen!60!white}38.38\%
     & \cellcolor{oliveGreen!48!white}47.08\%
     & \cellcolor{oliveGreen!62!white}\textbf{38.33\%}
     & \cellcolor{oliveGreen!90!white}61.17\%
  & 0.3 & \cellcolor{oliveGreen!88!white}90.67\%
     & \cellcolor{oliveGreen!87!white}92.00\%
     & \cellcolor{oliveGreen!65!white}37.37\%
     & \cellcolor{oliveGreen!81!white}45.42\%
     & \cellcolor{oliveGreen!85!white}37.92\%
     & \cellcolor{oliveGreen!76!white}60.68\% \\

0.4  & \cellcolor{oliveGreen!37!white}89.61\%
     & \cellcolor{oliveGreen!86!white}92.80\%
     & \cellcolor{oliveGreen!85!white}40.91\%
     & \cellcolor{oliveGreen!90!white}48.33\%
     & \cellcolor{oliveGreen!20!white}32.50\%
     & \cellcolor{oliveGreen!85!white}60.83\%
  & 0.4 & \cellcolor{oliveGreen!90!white}\textbf{90.98\%}
     & \cellcolor{oliveGreen!86!white}91.40\%
     & \cellcolor{oliveGreen!68!white}37.88\%
     & \cellcolor{oliveGreen!85!white}47.92\%
     & \cellcolor{oliveGreen!83!white}36.67\%
     & \cellcolor{oliveGreen!85!white}60.97\% \\

0.5  & \cellcolor{oliveGreen!73!white}90.45\%
     & \cellcolor{oliveGreen!90!white}\textbf{93.00\%}
     & \cellcolor{oliveGreen!20!white}34.34\%
     & \cellcolor{oliveGreen!71!white}\textbf{50.83\%}
     & \cellcolor{oliveGreen!62!white}36.25\%
     & \cellcolor{oliveGreen!66!white}60.97\%
  & 0.5 & \cellcolor{oliveGreen!85!white}90.37\%
     & \cellcolor{oliveGreen!90!white}91.20\%
     & \cellcolor{oliveGreen!90!white}\textbf{42.42\%}
     & \cellcolor{oliveGreen!85!white}47.92\%
     & \cellcolor{oliveGreen!83!white}35.83\%
     & \cellcolor{oliveGreen!86!white}61.55\% \\

0.6  & \cellcolor{oliveGreen!90!white}\textbf{90.83\%}
     & \cellcolor{oliveGreen!68!white}92.00\%
     & \cellcolor{oliveGreen!70!white}39.39\%
     & \cellcolor{oliveGreen!20!white}44.58\%
     & \cellcolor{oliveGreen!68!white}37.92\%
     & \cellcolor{oliveGreen!65!white}60.94\%
  & 0.6 & \cellcolor{oliveGreen!87!white}90.59\%
     & \cellcolor{oliveGreen!83!white}90.40\%
     & \cellcolor{oliveGreen!90!white}\textbf{42.42\%}
     & \cellcolor{oliveGreen!76!white}42.50\%
     & \cellcolor{oliveGreen!85!white}36.67\%
     & \cellcolor{oliveGreen!83!white}60.52\% \\

0.7  & \cellcolor{oliveGreen!56!white}90.06\%
     & \cellcolor{oliveGreen!59!white}91.60\%
     & \cellcolor{oliveGreen!50!white}37.37\%
     & \cellcolor{oliveGreen!25!white}45.00\%
     & \cellcolor{oliveGreen!59!white}37.08\%
     & \cellcolor{oliveGreen!45!white}60.22\%
  & \textcolor{oliveGreen}{\textbf{0.7}} \textcolor{oliveGreen}{\cmark} & \cellcolor{oliveGreen!89!white}90.83\%
     & \cellcolor{oliveGreen!90!white}\textbf{93.00\%}
     & \cellcolor{oliveGreen!85!white}41.41\%
     & \cellcolor{oliveGreen!90!white}\textbf{50.83\%}
     & \cellcolor{oliveGreen!89!white}38.33\%
     & \cellcolor{oliveGreen!90!white}\textbf{62.88\%} \\

0.8  & \cellcolor{oliveGreen!80!white}90.60\%
     & \cellcolor{oliveGreen!68!white}92.00\%
     & \cellcolor{oliveGreen!50!white}37.37\%
     & \cellcolor{oliveGreen!62!white}48.33\%
     & \cellcolor{oliveGreen!55!white}35.42\%
     & \cellcolor{oliveGreen!59!white}60.74\%
  & 0.8 & \cellcolor{oliveGreen!83!white}89.99\%
     & \cellcolor{oliveGreen!88!white}92.20\%
     & \cellcolor{oliveGreen!75!white}39.39\%
     & \cellcolor{oliveGreen!87!white}49.17\%
     & \cellcolor{oliveGreen!83!white}35.83\%
     & \cellcolor{oliveGreen!85!white}61.32\% \\

0.9  & \cellcolor{oliveGreen!70!white}90.37\%
     & \cellcolor{oliveGreen!42!white}90.80\%
     & \cellcolor{oliveGreen!70!white}39.39\%
     & \cellcolor{oliveGreen!85!white}50.42\%
     & \cellcolor{oliveGreen!60!white}35.83\%
     & \cellcolor{oliveGreen!76!white}61.36\%
  & 0.9 & \cellcolor{oliveGreen!84!white}90.22\%
     & \cellcolor{oliveGreen!88!white}92.20\%
     & \cellcolor{oliveGreen!83!white}40.91\%
     & \cellcolor{oliveGreen!87!white}48.75\%
     & \cellcolor{oliveGreen!75!white}32.50\%
     & \cellcolor{oliveGreen!83!white}60.52\% \\

\textcolor{oliveGreen}{\textbf{1.0}} \textcolor{oliveGreen}{\cmark} & \cellcolor{oliveGreen!60!white}90.14\%
     & \cellcolor{oliveGreen!37!white}90.60\%
     & \cellcolor{oliveGreen!90!white}\textbf{41.41\%}
     & \cellcolor{oliveGreen!71!white}49.17\%
     & \cellcolor{oliveGreen!85!white}37.92\%
     & \cellcolor{oliveGreen!90!white}\textbf{61.85\%}
  & 1.0 & \cellcolor{oliveGreen!86!white}90.44\%
     & \cellcolor{oliveGreen!85!white}91.00\%
     & \cellcolor{oliveGreen!45!white}33.33\%
     & \cellcolor{oliveGreen!83!white}46.67\%
     & \cellcolor{oliveGreen!90!white}\textbf{38.75\%}
     & \cellcolor{oliveGreen!82!white}60.04\% \\
\bottomrule
\end{tabular}
\end{table}

\swireasoning uses dwell windows (Sec.~\ref{sec:window}) to enforce the model stays in a thinking block for at least $W$ steps before switching to the other thinking mode. We conduct ablation studies on Qwen3-1.7B with a representative setting consisting of $W_{\mathrm{E}\to\mathrm{L}}\!\in\!\{64,128,256,512,1024\}$ and report Pass@1 accuracy on five benchmarks. Results in Tab.~\ref{tab:ablation-window} demonstrate that an intermediate window size of 512 consistently produces the best results.

When window sizes are too small, LLMs may jump back to latent mode prematurely, before explicit reasoning has consolidated a coherent trajectory. This increases exposure to noisy signals and harms final accuracy, especially on difficult tasks such as AIME24 and AIME25. When window sizes are too large, LLMs become sluggish to reenter latent exploration as confidence declines. A promising improvement direction is to make $W$ adaptive to the model's real-time density of effective reasoning.

\paragraph{Thinking-Related Signal Mixing.}

\swireasoning uses $\alpha_0,\beta_0\!\in\![0,1]$ as the initial ratios for mixing thinking-related signals at switching instants (Sec.~\ref{sec:mixing}). We sweep $\alpha_0$ and $\beta_0$ independently and report Pass@1 accuracies in Tab.~\ref{tab:ablation-alpha-beta}.

For the exit bias $\beta_0$, a very small $\beta_0$ implies excessive interference with when to conclude the thinking process and severely degrades accuracy (\textit{e.g.}, AIME24 drops to 8.33\% at $\beta_0{=}0.0$). Performance rises sharply and peaks near $\beta_0{=}0.7$, which achieves the best average 62.88\% and is either the best or the second-best on most datasets. A promising improvement direction is to make $\beta_0$ difficulty-aware, so that it will be automatically adjusted based on problem difficulty.

The situation for the entrance bias $\alpha_0$ is different. We observe a broad performance plateau for $\alpha_0\!\in\![0.4,0.9]$, with the highest average at $\alpha_0{=}1.0$ (61.85\%), however, only marginally higher than other values like $\alpha_0{=}0.9$ (61.36\%). Task-wise, problems with different levels of difficulty tend to have various preferences over $\alpha_0$. We expose $\alpha_0$ to users for adjustment based on task difficulty. The more detailed hyperparameters we adopted for the experiments are provided in Appendix~\ref{supp_hyperparameter}.

\begin{wraptable}{r}{0.5\textwidth}  
\scriptsize
\centering
\captionsetup{font={small}}
\caption{Ablation on switch window size. \textcolor{oliveGreen}{\textbf{Greener}} indicates better performance within each column.}
\label{tab:ablation-window}
\setlength{\tabcolsep}{2.2pt}
\begin{tabular}{lcccccc}
\toprule
\multirow{2}{*}[3pt]{\textbf{\makecell[l]{Window\\Size}}} &
\makecell{GSM8K} & 
\makecell{MATH \\ 500} & 
\makecell{GPQA \\ Diamond} & 
\makecell{AIME \\ 2024} & 
\makecell{AIME \\ 2025} &
\makecell{\textbf{Average}} \\
\cmidrule(r){1-1}\cmidrule{2-6}\cmidrule(l){7-7}
64   & \cellcolor{oliveGreen!20!white}89.69\%  
     & \cellcolor{oliveGreen!77!white}92.60\% 
     & \cellcolor{oliveGreen!76!white}40.91\% 
     & \cellcolor{oliveGreen!20!white}47.92\% 
     & \cellcolor{oliveGreen!20!white}34.17\% 
     & \cellcolor{oliveGreen!23!white}61.06\% \\
128  & \cellcolor{oliveGreen!67!white}90.45\%  
     & \cellcolor{oliveGreen!26!white}91.00\% 
     & \cellcolor{oliveGreen!20!white}38.89\% 
     & \cellcolor{oliveGreen!30!white}48.33\% 
     & \cellcolor{oliveGreen!55!white}36.25\% 
     & \cellcolor{oliveGreen!20!white}60.98\% \\
256  & \cellcolor{oliveGreen!24!white}89.76\%  
     & \cellcolor{oliveGreen!20!white}90.80\% 
     & \cellcolor{oliveGreen!48!white}39.90\% 
     & \cellcolor{oliveGreen!60!white}49.58\% 
     & \cellcolor{oliveGreen!55!white}36.25\% 
     & \cellcolor{oliveGreen!30!white}61.26\% \\
\textcolor{oliveGreen}{\textbf{512}} \textcolor{oliveGreen}{\cmark} & \cellcolor{oliveGreen!90!white}\textbf{90.83\%}  
     & \cellcolor{oliveGreen!90!white}\textbf{93.00\%} 
     & \cellcolor{oliveGreen!90!white}\textbf{41.41\%} 
     & \cellcolor{oliveGreen!90!white}\textbf{50.83\%} 
     & \cellcolor{oliveGreen!90!white}\textbf{38.33\%} 
     & \cellcolor{oliveGreen!90!white}\textbf{62.88\%} \\
1024 & \cellcolor{oliveGreen!90!white}\textbf{90.83\%}  
     & \cellcolor{oliveGreen!33!white}91.20\%
     & \cellcolor{oliveGreen!62!white}40.40\%
     & \cellcolor{oliveGreen!60!white}49.58\%
     & \cellcolor{oliveGreen!62!white}36.67\%
     & \cellcolor{oliveGreen!48!white}61.74\% \\
\bottomrule
\end{tabular}
\end{wraptable}

\paragraph{Maximum Switch Count.}

We suppress overthinking by bounding the number of mode switches with a budget $C_{\max}$ (Sec.~\ref{sec:switch_control}), and reducing $C_{\max}$ leads to earlier convergence. In Fig.~\ref{fig:eff} and Fig.~\ref{fig:eff-single}, moving rightward on the $x$–axis corresponds to smaller token budgets, \textit{i.e.}, smaller $C_{\max}$. We collect data points in these figures by incrementing the value of $C_{\max}$ from 1 until further increases in $C_{\max}$ no longer alter generation results in most cases, \textit{i.e.}, maximum accuracy is reached at saturation. Detailed data is provided in Appendix \ref{supp_exp_eff}.

As analyzed in Sec.~\ref{sec:exp_efficiency}, decreasing $C_{\max}$ yields a significant improvement in token efficiency, which confirms the intended behavior of the switch count control design: it curbs prolonged latent exploration and commits to an answer path early, thereby mitigating overthinking. With switch count control, a small number of confidence-aware blocks usually suffices for easy-to-moderate problems, while difficult instances benefit more from allowing a few more switches before the final answer. 

More ablation studies are provided in Appendix~\ref{abl:triggers}--\ref{abl:component}.

\subsection{Experimental Results on Larger Models}

\begin{table*}[ht]
\small
\centering
\captionsetup{font={small}}
\setlength{\tabcolsep}{0.6mm}
\caption{Comparison of \swireasoning and CoT with sampling, CoT with greedy decoding, and Soft Thinking on Qwen3-32B. \swireasoning improves accuracy by \textcolor{oliveGreen}{$\boldsymbol{+1.92\%}$} on average.}
\label{tab:accuracy_qwen3_32b}

\begin{tabular}{l r l  r l  r l  r l  r l  r l}
\toprule
\multirow{2}{*}[5pt]{\makecell{\textbf{Method}}} &
\multicolumn{2}{c}{\makecell{GSM8K}} &
\multicolumn{2}{c}{\makecell{MATH\\500}} &
\multicolumn{2}{c}{\makecell{GPQA\\Diamond}} &
\multicolumn{2}{c}{\makecell{AIME\\2024}} &
\multicolumn{2}{c}{\makecell{AIME\\2025}} &
\multicolumn{2}{c}{\makecell{\textbf{Average}}} \\
\cmidrule(r){1-1}\cmidrule(r){2-11}\cmidrule{12-13} 
CoT               & 95.83 & 
                  & 97.40 & 
                  & 66.16 & 
                  & 80.42 & 
                  & 72.08 & 
                  & 82.38 & \multicolumn{1}{c}{\scriptsize\textcolor{gray}{$+0.00$}} \\
CoT (Greedy)      & 95.91 & 
                  & 97.20 & 
                  & 69.70 & 
                  & 80.00 & 
                  & 73.33 & 
                  & 83.23 & \multicolumn{1}{c}{\scriptsize\textcolor{oliveGreen}{$+0.85$}} \\
Soft Thinking     & 95.75 & 
                  & 97.40 & 
                  & 67.17 & 
                  & 74.58 & 
                  & 66.25 & 
                  & 80.23 & \multicolumn{1}{c}{\scriptsize\textcolor{mauve}{$-2.15$}} \\
\rowcolor{gray!20}\textit{\textbf{SwiR} (Ours)}       
                  & \textbf{96.21} & \Deltabar{ 0.38}
                  & \textbf{98.40} & \Deltabar{ 1.00}
                  & \textbf{70.20} & \Deltabar{ 4.04}
                  & \textbf{82.92} & \Deltabar{ 2.50}
                  & \textbf{73.75} & \Deltabar{ 1.67}
                  & \textbf{84.30} & \Deltabar{ 1.92} \\
\bottomrule
\end{tabular}
\end{table*}

In addition to the 1.7B and 8B settings, we evaluate \swireasoning on the 32B scale using Qwen3-32B. We report pass@1 accuracy on GSM8K, MATH500, GPQA Diamond, AIME 2024, and AIME 2025. Baseline hyperparameters follow the recommendations from their original papers, and all methods use the same configuration as in the main paper. Table~\ref{tab:accuracy_qwen3_32b} shows that under no token budget constraint, \swireasoning improves accuracy by $+1.92\%$ on average over standard CoT. The gains are most notable on more difficult benchmarks such as GPQA Diamond ($+4.04\%$). These results indicate that \swireasoning scales to larger models and achieves consistent accuracy gains.

\subsection{Experimental Results on Broader Domains}

\begin{table*}[ht]
\small
\centering
\captionsetup{font={small}}
\setlength{\tabcolsep}{3pt}
\renewcommand{\BARMAX}{20}
\caption{Comparison of \swireasoning and baselines on coding, multi-hop QA, and commonsense reasoning tasks. \swireasoning improves accuracy by \textcolor{oliveGreen}{$\boldsymbol{+2.70\%}$} on average.}
\label{tab:broader_domains}
\vspace{-0.5em}
\begin{tabular}{l r l  r l  r l  r l  r l}
\toprule
\multirow{2}{*}[-2pt]{\makecell{\textbf{Method}}} &
\multicolumn{2}{c}{\multirow{2}{*}[-2pt]{\makecell{HumanEval}}} &
\multicolumn{6}{c}{\makecell{LeetCode-Contest}} &
\multicolumn{2}{c}{\multirow{2}{*}[-2pt]{\makecell{MBPP}}} \\
\cmidrule(lr){4-9}
& & &
\multicolumn{2}{c}{\textit{Easy-level}} &
\multicolumn{2}{c}{\textit{Medium-level}} &
\multicolumn{2}{c}{\textit{Hard-level}} &
& \\
\midrule
CoT               & 92.68 &
                  & 57.78 &
                  & 68.13 &
                  & 43.18 &
                  & 94.16 &
                  \\
CoT (Greedy)      & 93.90 &
                  & 64.44 &
                  & 58.24 &
                  & 47.73 &
                  & 91.44 &
                  \\
Soft Thinking     & 92.07 &
                  & 55.56 &
                  & 61.54 &
                  & 38.64 &
                  & 94.16 &
                  \\
\rowcolor{gray!20}\textit{\textbf{SwiR} (Ours)} &
                  \textbf{95.73} & \Deltabar{ 3.05}
                  & \textbf{64.44} & \Deltabar{ 6.66}
                  & \textbf{69.23} & \Deltabar{ 1.10}
                  & \textbf{61.36} & \Deltabar{18.18}
                  & \textbf{95.33} & \Deltabar{ 1.17} \\
\bottomrule
\end{tabular}

\vspace{8pt}

\begin{tabular}{l r l  r l  r l  r l}
\toprule
\textbf{Method} &
\multicolumn{2}{c}{\makecell{LiveCode-\\Bench}} &
\multicolumn{2}{c}{\makecell{2WikiMul-\\tihopQA}} &
\multicolumn{2}{c}{\makecell{Common-\\senseQA}} &
\multicolumn{2}{c}{\makecell{\textbf{Average}}} \\
\midrule
CoT               & 62.01 &
                  & 79.00 &
                  & 83.95 &
                  & 78.54 & \multicolumn{1}{c}{\scriptsize\textcolor{gray}{$+0.00$}} \\
CoT (Greedy)      & 50.18 &
                  & 79.50 &
                  & 83.95 &
                  & 76.03 & \multicolumn{1}{c}{\scriptsize\textcolor{mauve}{$-2.51$}} \\
Soft Thinking     & 56.99 &
                  & 79.00 &
                  & 83.70 &
                  & 76.73 & \multicolumn{1}{c}{\scriptsize\textcolor{mauve}{$-1.81$}} \\
\rowcolor{gray!20}\textit{\textbf{SwiR} (Ours)} &
                  \textbf{63.44} & \Deltabar{ 1.43}
                  & \textbf{81.50} & \Deltabar{ 2.50}
                  & \textbf{85.34} & \Deltabar{ 1.39}
                  & \textbf{81.24} & \Deltabar{ 2.70} \\
\bottomrule
\end{tabular}
\end{table*}

In addition to math and STEM domains, we further evaluate \swireasoning on coding, multi-hop QA, and commonsense reasoning. We use Qwen3-8B and report pass@1 accuracy on HumanEval~\citep{chen2021evaluating}, LeetCode-Contest~\citep{guo2024deepseekcoder}, MBPP~\citep{austin2021program}, and LiveCodeBench~\citep{jain2024livecodebench} for coding, 2WikiMultihopQA~\citep{ho2020constructing} set from LongBench~\citep{bai2024longbench} for multi-hop QA, and CommonsenseQA~\citep{talmor2019commonsenseqa} for commonsense reasoning. As shown in Tab.~\ref {tab:broader_domains}, under no token budget constraint, \swireasoning improves accuracy by $+2.70\%$ on average over standard CoT (the reported average is the simple mean of accuracy over the full LeetCode-Contest and accuracies over five other benchmarks). 

On coding tasks, the largest gains ($+18.18\%$) are observed on the hard-level subset, indicating that \swireasoning is most helpful for problems that require stronger reasoning capabilities. On multi-hop QA tasks, which require retrieving and connecting multiple disparate facts, \swireasoning outperforms CoT by $+2.50\%$. This suggests that the exploration capability of latent reasoning is effective in navigating complex reasoning paths. On commonsense reasoning tasks, \swireasoning surpasses CoT by $+1.39\%$. This demonstrates the robustness of \swireasoning in general knowledge scenarios. Overall, these results confirm that \swireasoning generalizes to broader domains with consistent accuracy gains.
\section{Conclusion}

This paper presents \swireasoning, a training-free inference framework that integrates explicit chain-of-thought thinking with latent thinking through an entropy trends–based controller. The framework is conceptually straightforward but empirically effective: when block-wise uncertainty decreases, we collapse to a single explicit path to consolidate progress. When uncertainty rises and has persisted for a minimal dwell window, we expand into latent space to explore more alternatives. Complementing this mode switch, a switch count controller caps the number of transitions, thereby curbing overthinking while preserving prediction quality. Together, these two mechanisms yield consistently improved Pareto frontiers for reasoning LLMs, effectively enhancing both maximum accuracy under unlimited budgets and token efficiency under limited budgets. Looking ahead, integrating \swireasoning with reinforcement learning–based training may unlock even stronger reasoning capabilities.
\newpage
\subsubsection*{Acknowledgments}

The authors would like to thank Microsoft and Georgia Tech for their support. This material is based upon work supported by the National Science Foundation under grant no. 2229876 and is supported in part by funds provided by the National Science Foundation, by the Department of Homeland Security, and by IBM. Any opinions, findings, and conclusions or recommendations expressed in this material are those of the author(s) and do not necessarily reflect the views of the National Science Foundation or its federal agency and industry partners.
\section{Ethics Statement}

This work focuses on enhancing the reasoning accuracy and token efficiency of LLMs, which does not raise safety concerns. This work involves no collection of sensitive data. All evaluations are conducted using publicly available models and benchmarks under their original licenses.

\section{Reproducibility Statement}

We provide implementation details in Appendix~\ref{supp_imp}, details of the benchmark settings in Appendix~\ref{supp_benchmark}, and details of the hyperparameters in Appendix~\ref{supp_hyperparameter} to facilitate reproducibility.

\bibliography{main}

\begin{thebibliography}{80}
\providecommand{\natexlab}[1]{#1}
\providecommand{\url}[1]{\texttt{#1}}
\expandafter\ifx\csname urlstyle\endcsname\relax
  \providecommand{\doi}[1]{doi: #1}\else
  \providecommand{\doi}{doi: \begingroup \urlstyle{rm}\Url}\fi

\bibitem[Abdin et~al.(2025)Abdin, Agarwal, Awadallah, Balachandran, Behl, Chen, de~Rosa, Gunasekar, Javaheripi, Joshi, et~al.]{abdin2025phi}
Marah Abdin, Sahaj Agarwal, Ahmed Awadallah, Vidhisha Balachandran, Harkirat Behl, Lingjiao Chen, Gustavo de~Rosa, Suriya Gunasekar, Mojan Javaheripi, Neel Joshi, et~al.
\newblock Phi-4-reasoning technical report.
\newblock \emph{arXiv preprint arXiv:2504.21318}, 2025.

\bibitem[Abouelenin et~al.(2025)Abouelenin, Ashfaq, Atkinson, Awadalla, Bach, Bao, Benhaim, Cai, Chaudhary, Chen, et~al.]{abouelenin2025phi}
Abdelrahman Abouelenin, Atabak Ashfaq, Adam Atkinson, Hany Awadalla, Nguyen Bach, Jianmin Bao, Alon Benhaim, Martin Cai, Vishrav Chaudhary, Congcong Chen, et~al.
\newblock Phi-4-mini technical report: Compact yet powerful multimodal language models via mixture-of-loras.
\newblock \emph{arXiv preprint arXiv:2503.01743}, 2025.

\bibitem[Agarwal et~al.(2025)Agarwal, Ahmad, Ai, Altman, Applebaum, Arbus, Arora, Bai, Baker, Bao, et~al.]{agarwal2025gpt}
Sandhini Agarwal, Lama Ahmad, Jason Ai, Sam Altman, Andy Applebaum, Edwin Arbus, Rahul~K Arora, Yu~Bai, Bowen Baker, Haiming Bao, et~al.
\newblock gpt-oss-120b \& gpt-oss-20b model card.
\newblock \emph{arXiv preprint arXiv:2508.10925}, 2025.

\bibitem[Anthropic(2025)]{anthropic2025claude4_system_card}
Anthropic.
\newblock System card: Claude opus 4 \& claude sonnet 4.
\newblock 2025.
\newblock URL \url{https://www.anthropic.com/claude-4-system-card}.

\bibitem[Austin et~al.(2021)Austin, Odena, Nye, Bosma, Michalewski, Dohan, Jiang, Cai, Terry, Le, et~al.]{austin2021program}
Jacob Austin, Augustus Odena, Maxwell Nye, Maarten Bosma, Henryk Michalewski, David Dohan, Ellen Jiang, Carrie Cai, Michael Terry, Quoc Le, et~al.
\newblock Program synthesis with large language models.
\newblock \emph{arXiv preprint arXiv:2108.07732}, 2021.

\bibitem[Bai et~al.(2024)Bai, Lv, Zhang, Lyu, Tang, Huang, Du, Liu, Zeng, Hou, et~al.]{bai2024longbench}
Yushi Bai, Xin Lv, Jiajie Zhang, Hongchang Lyu, Jiankai Tang, Zhidian Huang, Zhengxiao Du, Xiao Liu, Aohan Zeng, Lei Hou, et~al.
\newblock Longbench: A bilingual, multitask benchmark for long context understanding.
\newblock In \emph{Proceedings of the 62nd Annual Meeting of the Association for Computational Linguistics (Volume 1: Long Papers)}, pp.\  3119--3137, 2024.

\bibitem[Cai et~al.(2024)Cai, Zhang, Gao, Liu, Li, Liu, Lu, Xiong, Dong, Hu, et~al.]{cai2024pyramidkv}
Zefan Cai, Yichi Zhang, Bofei Gao, Yuliang Liu, Yucheng Li, Tianyu Liu, Keming Lu, Wayne Xiong, Yue Dong, Junjie Hu, et~al.
\newblock Pyramidkv: Dynamic kv cache compression based on pyramidal information funneling.
\newblock \emph{arXiv preprint arXiv:2406.02069}, 2024.

\bibitem[Chen et~al.(2021)Chen, Tworek, Jun, Yuan, de~Oliveira~Pinto, Kaplan, Edwards, Burda, Joseph, Brockman, Ray, Puri, Krueger, Petrov, Khlaaf, Sastry, Mishkin, Chan, Gray, Ryder, Pavlov, Power, Kaiser, Bavarian, Winter, Tillet, Such, Cummings, Plappert, Chantzis, Barnes, Herbert-Voss, Guss, Nichol, Paino, Tezak, Tang, Babuschkin, Balaji, Jain, Saunders, Hesse, Carr, Leike, Achiam, Misra, Morikawa, Radford, Knight, Brundage, Murati, Mayer, Welinder, McGrew, Amodei, McCandlish, Sutskever, and Zaremba]{chen2021evaluating}
Mark Chen, Jerry Tworek, Heewoo Jun, Qiming Yuan, Henrique~Ponde de~Oliveira~Pinto, Jared Kaplan, Harri Edwards, Yuri Burda, Nicholas Joseph, Greg Brockman, Alex Ray, Raul Puri, Gretchen Krueger, Michael Petrov, Heidy Khlaaf, Girish Sastry, Pamela Mishkin, Brooke Chan, Scott Gray, Nick Ryder, Mikhail Pavlov, Alethea Power, Lukasz Kaiser, Mohammad Bavarian, Clemens Winter, Philippe Tillet, Felipe~Petroski Such, Dave Cummings, Matthias Plappert, Fotios Chantzis, Elizabeth Barnes, Ariel Herbert-Voss, William~Hebgen Guss, Alex Nichol, Alex Paino, Nikolas Tezak, Jie Tang, Igor Babuschkin, Suchir Balaji, Shantanu Jain, William Saunders, Christopher Hesse, Andrew~N. Carr, Jan Leike, Josh Achiam, Vedant Misra, Evan Morikawa, Alec Radford, Matthew Knight, Miles Brundage, Mira Murati, Katie Mayer, Peter Welinder, Bob McGrew, Dario Amodei, Sam McCandlish, Ilya Sutskever, and Wojciech Zaremba.
\newblock Evaluating large language models trained on code, 2021.

\bibitem[Chen et~al.(2025)Chen, Zhao, Xia, Lu, Wang, Chen, Zhang, Wang, Li, and Shen]{chen2025reasoning}
Xinghao Chen, Anhao Zhao, Heming Xia, Xuan Lu, Hanlin Wang, Yanjun Chen, Wei Zhang, Jian Wang, Wenjie Li, and Xiaoyu Shen.
\newblock Reasoning beyond language: A comprehensive survey on latent chain-of-thought reasoning.
\newblock \emph{arXiv preprint arXiv:2505.16782}, 2025.

\bibitem[Chen et~al.(2024)Chen, May, Svirschevski, Huang, Ryabinin, Jia, and Chen]{chen2024sequoia}
Zhuoming Chen, Avner May, Ruslan Svirschevski, Yuhsun Huang, Max Ryabinin, Zhihao Jia, and Beidi Chen.
\newblock Sequoia: Scalable, robust, and hardware-aware speculative decoding.
\newblock \emph{arXiv preprint arXiv:2402.12374}, 2024.

\bibitem[Cheng \& Van~Durme(2024)Cheng and Van~Durme]{cheng2024compressed}
Jeffrey Cheng and Benjamin Van~Durme.
\newblock Compressed chain of thought: Efficient reasoning through dense representations.
\newblock \emph{arXiv preprint arXiv:2412.13171}, 2024.

\bibitem[Cobbe et~al.(2021)Cobbe, Kosaraju, Bavarian, Chen, Jun, Kaiser, Plappert, Tworek, Hilton, Nakano, et~al.]{cobbe2021training}
Karl Cobbe, Vineet Kosaraju, Mohammad Bavarian, Mark Chen, Heewoo Jun, Lukasz Kaiser, Matthias Plappert, Jerry Tworek, Jacob Hilton, Reiichiro Nakano, et~al.
\newblock Training verifiers to solve math word problems.
\newblock \emph{arXiv preprint arXiv:2110.14168}, 2021.

\bibitem[Dao(2023)]{dao2023flashattention}
Tri Dao.
\newblock Flashattention-2: Faster attention with better parallelism and work partitioning.
\newblock \emph{arXiv preprint arXiv:2307.08691}, 2023.

\bibitem[Dao et~al.(2022)Dao, Fu, Ermon, Rudra, and R{\'e}]{dao2022flashattention}
Tri Dao, Dan Fu, Stefano Ermon, Atri Rudra, and Christopher R{\'e}.
\newblock Flashattention: Fast and memory-efficient exact attention with io-awareness.
\newblock \emph{Advances in Neural Information Processing Systems}, 35:\penalty0 16344--16359, 2022.

\bibitem[{DeepMind}(2024{\natexlab{a}})]{deepmind2023alphazero}
{DeepMind}.
\newblock Alphazero: Shedding new light on chess, shogi, and go.
\newblock \emph{URL https://deepmind.google/discover/blog/alphazero-shedding-new-light-on-chess-shogi-and-go/}, 2024{\natexlab{a}}.

\bibitem[{DeepMind}(2024{\natexlab{b}})]{deepmind2023imo}
{DeepMind}.
\newblock Ai solves imo problems at a silver medal level.
\newblock \emph{URL https://deepmind.google/discover/blog/ai-solves-imo-problems-at-silver-medal-level/}, 2024{\natexlab{b}}.

\bibitem[Fu et~al.(2024)Fu, Yu, Li, Qian, Zhang, Yuan, Shi, Yakunin, and Lin]{fu2024amoeballm}
Yonggan Fu, Zhongzhi Yu, Junwei Li, Jiayi Qian, Yongan Zhang, Xiangchi Yuan, Dachuan Shi, Roman Yakunin, and Yingyan~Celine Lin.
\newblock Amoeballm: Constructing any-shape large language models for efficient and instant deployment.
\newblock \emph{Advances in Neural Information Processing Systems}, 37:\penalty0 78299--78319, 2024.

\bibitem[Goyal et~al.(2024)Goyal, Ji, Rawat, Menon, Kumar, and Nagarajan]{goyalthink}
Sachin Goyal, Ziwei Ji, Ankit~Singh Rawat, Aditya~Krishna Menon, Sanjiv Kumar, and Vaishnavh Nagarajan.
\newblock Think before you speak: Training language models with pause tokens.
\newblock In \emph{The Twelfth International Conference on Learning Representations}, 2024.

\bibitem[Guo et~al.(2024)Guo, Zhu, Yang, Xie, Dong, Zhang, Chen, Bi, Wu, Li, Luo, Xiong, and Liang]{guo2024deepseekcoder}
Daya Guo, Qihao Zhu, Dejian Yang, Zhenda Xie, Kai Dong, Wentao Zhang, Guanting Chen, Xiao Bi, Y.~Wu, Y.~K. Li, Fuli Luo, Yingfei Xiong, and Wenfeng Liang.
\newblock Deepseek-coder: When the large language model meets programming - the rise of code intelligence.
\newblock \emph{arXiv preprint arXiv: 2401.14196}, 2024.

\bibitem[Guo et~al.(2025)Guo, Yang, Zhang, Song, Zhang, Xu, Zhu, Ma, Wang, Bi, et~al.]{guo2025deepseek}
Daya Guo, Dejian Yang, Haowei Zhang, Junxiao Song, Ruoyu Zhang, Runxin Xu, Qihao Zhu, Shirong Ma, Peiyi Wang, Xiao Bi, et~al.
\newblock Deepseek-r1: Incentivizing reasoning capability in llms via reinforcement learning.
\newblock \emph{arXiv preprint arXiv:2501.12948}, 2025.

\bibitem[Han et~al.(2023)Han, Wang, Peng, Xiong, Chen, Ji, and Wang]{han2023lm}
Chi Han, Qifan Wang, Hao Peng, Wenhan Xiong, Yu~Chen, Heng Ji, and Sinong Wang.
\newblock Lm-infinite: Zero-shot extreme length generalization for large language models.
\newblock \emph{arXiv preprint arXiv:2308.16137}, 2023.

\bibitem[Hao et~al.(2024)Hao, Sukhbaatar, Su, Li, Hu, Weston, and Tian]{hao2024training}
Shibo Hao, Sainbayar Sukhbaatar, DiJia Su, Xian Li, Zhiting Hu, Jason Weston, and Yuandong Tian.
\newblock Training large language models to reason in a continuous latent space.
\newblock \emph{arXiv preprint arXiv:2412.06769}, 2024.

\bibitem[Hendrycks et~al.(2021)Hendrycks, Burns, Kadavath, Arora, Basart, Tang, Song, and Steinhardt]{hendrycks2021measuring}
Dan Hendrycks, Collin Burns, Saurav Kadavath, Akul Arora, Steven Basart, Eric Tang, Dawn Song, and Jacob Steinhardt.
\newblock Measuring mathematical problem solving with the math dataset.
\newblock \emph{arXiv preprint arXiv:2103.03874}, 2021.

\bibitem[Ho et~al.(2020)Ho, Nguyen, Sugawara, and Aizawa]{ho2020constructing}
Xanh Ho, Anh-Khoa~Duong Nguyen, Saku Sugawara, and Akiko Aizawa.
\newblock Constructing a multi-hop qa dataset for comprehensive evaluation of reasoning steps.
\newblock \emph{arXiv preprint arXiv:2011.01060}, 2020.

\bibitem[HuggingFaceH4(2024)]{hf_aime24_dataset}
HuggingFaceH4.
\newblock Aime 2024 (american invitational mathematics examination 2024).
\newblock Hugging Face dataset, 2024.
\newblock URL \url{https://huggingface.co/datasets/HuggingFaceH4/aime_2024}.

\bibitem[Jaech et~al.(2024)Jaech, Kalai, Lerer, Richardson, El-Kishky, Low, Helyar, Madry, Beutel, Carney, et~al.]{jaech2024openai}
Aaron Jaech, Adam Kalai, Adam Lerer, Adam Richardson, Ahmed El-Kishky, Aiden Low, Alec Helyar, Aleksander Madry, Alex Beutel, Alex Carney, et~al.
\newblock Openai o1 system card.
\newblock \emph{arXiv preprint arXiv:2412.16720}, 2024.

\bibitem[Jain et~al.(2024)Jain, Han, Gu, Li, Yan, Zhang, Wang, Solar-Lezama, Sen, and Stoica]{jain2024livecodebench}
Naman Jain, King Han, Alex Gu, Wen-Ding Li, Fanjia Yan, Tianjun Zhang, Sida Wang, Armando Solar-Lezama, Koushik Sen, and Ion Stoica.
\newblock Livecodebench: Holistic and contamination free evaluation of large language models for code.
\newblock \emph{arXiv preprint arXiv:2403.07974}, 2024.

\bibitem[Jiang et~al.(2023)Jiang, Wu, Lin, Yang, and Qiu]{jiang2023llmlingua}
Huiqiang Jiang, Qianhui Wu, Chin-Yew Lin, Yuqing Yang, and Lili Qiu.
\newblock Llmlingua: Compressing prompts for accelerated inference of large language models.
\newblock \emph{arXiv preprint arXiv:2310.05736}, 2023.

\bibitem[Jiang et~al.(2025)Jiang, Wu, Zhan, Lai, and Lian]{jiang2025}
Nan Jiang, Ziming Wu, De-Chuan Zhan, Fuming Lai, and Shaobing Lian.
\newblock Dart: Distilling autoregressive reasoning to silent thought, 2025.

\bibitem[Kim et~al.(2023)Kim, Mangalam, Moon, Malik, Mahoney, Gholami, and Keutzer]{kim2023speculative}
Sehoon Kim, Karttikeya Mangalam, Suhong Moon, Jitendra Malik, Michael~W Mahoney, Amir Gholami, and Kurt Keutzer.
\newblock Speculative decoding with big little decoder.
\newblock \emph{Advances in Neural Information Processing Systems}, 36:\penalty0 39236--39256, 2023.

\bibitem[Kojima et~al.(2022)Kojima, Gu, Reid, Matsuo, and Iwasawa]{kojima2022large}
Takeshi Kojima, Shixiang~Shane Gu, Machel Reid, Yutaka Matsuo, and Yusuke Iwasawa.
\newblock Large language models are zero-shot reasoners.
\newblock \emph{Advances in neural information processing systems}, 35:\penalty0 22199--22213, 2022.

\bibitem[Leviathan et~al.(2023)Leviathan, Kalman, and Matias]{leviathan2023fast}
Yaniv Leviathan, Matan Kalman, and Yossi Matias.
\newblock Fast inference from transformers via speculative decoding.
\newblock In \emph{International Conference on Machine Learning}, pp.\  19274--19286. PMLR, 2023.

\bibitem[Li et~al.(2025{\natexlab{a}})Li, Li, Wu, Zhu, Wang, Yu, Jiang, Zhu, Jia, Wu, et~al.]{li2025seek}
Hengli Li, Chenxi Li, Tong Wu, Xuekai Zhu, Yuxuan Wang, Zhaoxin Yu, Eric~Hanchen Jiang, Song-Chun Zhu, Zixia Jia, Ying~Nian Wu, et~al.
\newblock Seek in the dark: Reasoning via test-time instance-level policy gradient in latent space.
\newblock \emph{arXiv preprint arXiv:2505.13308}, 2025{\natexlab{a}}.

\bibitem[Li et~al.(2025{\natexlab{b}})Li, Fu, Fan, Liu, Shu, Qin, Yang, King, and Ying]{li2025implicit}
Jindong Li, Yali Fu, Li~Fan, Jiahong Liu, Yao Shu, Chengwei Qin, Menglin Yang, Irwin King, and Rex Ying.
\newblock Implicit reasoning in large language models: A comprehensive survey.
\newblock \emph{arXiv preprint arXiv:2509.02350}, 2025{\natexlab{b}}.

\bibitem[Lin et~al.(2024)Lin, Tang, Tang, Yang, Chen, Wang, Xiao, Dang, Gan, and Han]{lin2024awq}
Ji~Lin, Jiaming Tang, Haotian Tang, Shang Yang, Wei-Ming Chen, Wei-Chen Wang, Guangxuan Xiao, Xingyu Dang, Chuang Gan, and Song Han.
\newblock Awq: Activation-aware weight quantization for on-device llm compression and acceleration.
\newblock \emph{Proceedings of machine learning and systems}, 6:\penalty0 87--100, 2024.

\bibitem[Liu et~al.(2024)Liu, Chen, Liu, Tian, and Luo]{liu2024expediting}
Tianqiao Liu, Zui Chen, Zitao Liu, Mi~Tian, and Weiqi Luo.
\newblock Expediting and elevating large language model reasoning via hidden chain-of-thought decoding.
\newblock \emph{arXiv preprint arXiv:2409.08561}, 2024.

\bibitem[Liu et~al.(2023)Liu, Hu, Bailis, Cheung, Deng, Stoica, and Zhang]{liu2023online}
Xiaoxuan Liu, Lanxiang Hu, Peter Bailis, Alvin Cheung, Zhijie Deng, Ion Stoica, and Hao Zhang.
\newblock Online speculative decoding.
\newblock \emph{arXiv preprint arXiv:2310.07177}, 2023.

\bibitem[Meta(2025{\natexlab{a}})]{meta2025llama}
AI~Meta.
\newblock The llama 4 herd: The beginning of a new era of natively multimodal ai innovation.
\newblock \emph{https://ai. meta. com/blog/llama-4-multimodal-intelligence/, checked on}, 4\penalty0 (7):\penalty0 2025, 2025{\natexlab{a}}.

\bibitem[Meta(2025{\natexlab{b}})]{meta_llama_llama3_3_model_card}
AI~Meta.
\newblock Llama 3.3 model card.
\newblock \url{https://github.com/meta-llama/llama-models/blob/main/models/llama3_3/MODEL_CARD.md}, 2025{\natexlab{b}}.

\bibitem[Ning et~al.(2025)Ning, Li, Fang, Tan, and Liu]{ning2025not}
Yansong Ning, Wei Li, Jun Fang, Naiqiang Tan, and Hao Liu.
\newblock Not all thoughts are generated equal: Efficient llm reasoning via multi-turn reinforcement learning.
\newblock \emph{arXiv preprint arXiv:2505.11827}, 2025.

\bibitem[OpenAI(2025{\natexlab{a}})]{openai2025gpt5systemcard}
OpenAI.
\newblock Gpt-5 system card.
\newblock System Card v2025-08-13, OpenAI, San Francisco, CA, August 2025{\natexlab{a}}.
\newblock URL \url{https://cdn.openai.com/gpt-5-system-card.pdf}.

\bibitem[OpenAI(2025{\natexlab{b}})]{openai2025o3}
OpenAI.
\newblock Openai o3-mini.
\newblock \emph{URL https://openai.com/index/openai-o3-mini/}, 2025{\natexlab{b}}.

\bibitem[Pan et~al.(2024)Pan, Wu, Jiang, Xia, Luo, Zhang, Lin, R{\"u}hle, Yang, Lin, et~al.]{pan2024llmlingua}
Zhuoshi Pan, Qianhui Wu, Huiqiang Jiang, Menglin Xia, Xufang Luo, Jue Zhang, Qingwei Lin, Victor R{\"u}hle, Yuqing Yang, Chin-Yew Lin, et~al.
\newblock Llmlingua-2: Data distillation for efficient and faithful task-agnostic prompt compression.
\newblock \emph{arXiv preprint arXiv:2403.12968}, 2024.

\bibitem[Pfau et~al.(2024)Pfau, Merrill, and Bowman]{pfau2024let}
Jacob Pfau, William Merrill, and Samuel~R Bowman.
\newblock Let's think dot by dot: Hidden computation in transformer language models.
\newblock \emph{arXiv preprint arXiv:2404.15758}, 2024.

\bibitem[{Qwen Team}(2024)]{qwen2.5}
{Qwen Team}.
\newblock Qwen2.5 technical report.
\newblock \emph{arXiv preprint arXiv:2412.15115}, 2024.

\bibitem[{Qwen Team}(2025)]{qwq32b}
{Qwen Team}.
\newblock Qwq-32b: Embracing the power of reinforcement learning, March 2025.
\newblock URL \url{https://qwenlm.github.io/blog/qwq-32b/}.

\bibitem[Rein et~al.(2024)Rein, Hou, Stickland, Petty, Pang, Dirani, Michael, and Bowman]{rein2024gpqa}
David Rein, Betty~Li Hou, Asa~Cooper Stickland, Jackson Petty, Richard~Yuanzhe Pang, Julien Dirani, Julian Michael, and Samuel~R Bowman.
\newblock Gpqa: A graduate-level google-proof q\&a benchmark.
\newblock In \emph{First Conference on Language Modeling}, 2024.

\bibitem[Shah et~al.(2024)Shah, Bikshandi, Zhang, Thakkar, Ramani, and Dao]{shah2024flashattention}
Jay Shah, Ganesh Bikshandi, Ying Zhang, Vijay Thakkar, Pradeep Ramani, and Tri Dao.
\newblock Flashattention-3: Fast and accurate attention with asynchrony and low-precision.
\newblock \emph{Advances in Neural Information Processing Systems}, 37:\penalty0 68658--68685, 2024.

\bibitem[Shen et~al.(2025)Shen, Yan, Zhang, Hu, Du, and He]{shen2025codi}
Zhenyi Shen, Hanqi Yan, Linhai Zhang, Zhanghao Hu, Yali Du, and Yulan He.
\newblock Codi: Compressing chain-of-thought into continuous space via self-distillation.
\newblock \emph{arXiv preprint arXiv:2502.21074}, 2025.

\bibitem[Shi et~al.(2023)Shi, Tao, Jin, Yang, Yuan, and Wang]{shi2023upop}
Dachuan Shi, Chaofan Tao, Ying Jin, Zhendong Yang, Chun Yuan, and Jiaqi Wang.
\newblock Upop: Unified and progressive pruning for compressing vision-language transformers.
\newblock In \emph{International Conference on Machine Learning}, pp.\  31292--31311. PMLR, 2023.

\bibitem[Shi et~al.(2024)Shi, Tao, Rao, Yang, Yuan, and Wang]{shi2023crossget}
Dachuan Shi, Chaofan Tao, Anyi Rao, Zhendong Yang, Chun Yuan, and Jiaqi Wang.
\newblock Crossget: Cross-guided ensemble of tokens for accelerating vision-language transformers.
\newblock In \emph{Forty-First International Conference on Machine Learning}, 2024.

\bibitem[Shi et~al.(2025)Shi, Fu, Yuan, Yu, You, Li, Dong, Kautz, Molchanov, and Lin]{shilacache}
Dachuan Shi, Yonggan Fu, Xiangchi Yuan, Zhongzhi Yu, Haoran You, Sixu Li, Xin Dong, Jan Kautz, Pavlo Molchanov, and Yingyan~Celine Lin.
\newblock Lacache: Ladder-shaped kv caching for efficient long-context modeling of large language models.
\newblock In \emph{Forty-second International Conference on Machine Learning}, 2025.

\bibitem[Su et~al.(2025)Su, Zhu, Xu, Jiao, Tian, and Zheng]{su2025token}
DiJia Su, Hanlin Zhu, Yingchen Xu, Jiantao Jiao, Yuandong Tian, and Qinqing Zheng.
\newblock Token assorted: Mixing latent and text tokens for improved language model reasoning.
\newblock \emph{arXiv preprint arXiv:2502.03275}, 2025.

\bibitem[Sun et~al.(2023)Sun, Suresh, Ro, Beirami, Jain, and Yu]{sun2023spectr}
Ziteng Sun, Ananda~Theertha Suresh, Jae~Hun Ro, Ahmad Beirami, Himanshu Jain, and Felix Yu.
\newblock Spectr: Fast speculative decoding via optimal transport.
\newblock \emph{Advances in Neural Information Processing Systems}, 36:\penalty0 30222--30242, 2023.

\bibitem[Tack et~al.(2025)Tack, Lanchantin, Yu, Cohen, Kulikov, Lan, Hao, Tian, Weston, and Li]{tack2025llm}
Jihoon Tack, Jack Lanchantin, Jane Yu, Andrew Cohen, Ilia Kulikov, Janice Lan, Shibo Hao, Yuandong Tian, Jason Weston, and Xian Li.
\newblock Llm pretraining with continuous concepts.
\newblock \emph{arXiv preprint arXiv:2502.08524}, 2025.

\bibitem[Talmor et~al.(2019)Talmor, Herzig, Lourie, and Berant]{talmor2019commonsenseqa}
Alon Talmor, Jonathan Herzig, Nicholas Lourie, and Jonathan Berant.
\newblock Commonsenseqa: A question answering challenge targeting commonsense knowledge.
\newblock In \emph{Proceedings of the 2019 Conference of the North American Chapter of the Association for Computational Linguistics: Human Language Technologies, Volume 1 (Long and Short Papers)}, pp.\  4149--4158, 2019.

\bibitem[Tan et~al.(2025)Tan, Li, Ju, Luo, Luan, and Song]{tan2025think}
Wenhui Tan, Jiaze Li, Jianzhong Ju, Zhenbo Luo, Jian Luan, and Ruihua Song.
\newblock Think silently, think fast: Dynamic latent compression of llm reasoning chains.
\newblock \emph{arXiv preprint arXiv:2505.16552}, 2025.

\bibitem[Wang et~al.(2025{\natexlab{a}})Wang, Wang, Zhu, and Liu]{wang2025system}
Xiaoqiang Wang, Suyuchen Wang, Yun Zhu, and Bang Liu.
\newblock System-1.5 reasoning: Traversal in language and latent spaces with dynamic shortcuts.
\newblock \emph{arXiv preprint arXiv:2505.18962}, 2025{\natexlab{a}}.

\bibitem[Wang et~al.(2025{\natexlab{b}})Wang, Wang, Ying, Bai, Gong, Dong, Liu, and Fu]{wang2025}
Xinyuan Wang, Dongjie Wang, Wangyang Ying, Haoyue Bai, Nanxu Gong, Sixun Dong, Kunpeng Liu, and Yanjie Fu.
\newblock Efficient post-training refinement of latent reasoning in large language models, 2025{\natexlab{b}}.

\bibitem[Wang et~al.(2022)Wang, Wei, Schuurmans, Le, Chi, Narang, Chowdhery, and Zhou]{wang2022self}
Xuezhi Wang, Jason Wei, Dale Schuurmans, Quoc Le, Ed~Chi, Sharan Narang, Aakanksha Chowdhery, and Denny Zhou.
\newblock Self-consistency improves chain of thought reasoning in language models.
\newblock \emph{arXiv preprint arXiv:2203.11171}, 2022.

\bibitem[Wei et~al.(2022)Wei, Wang, Schuurmans, Bosma, and et~al.]{wei2022chain}
Jason Wei, Xuezhi Wang, Dale Schuurmans, Maarten Bosma, and et~al.
\newblock Chain-of-thought prompting elicits reasoning in large language models.
\newblock \emph{arXiv preprint arXiv:2201.11903}, 2022.

\bibitem[Wingate et~al.(2022)Wingate, Shoeybi, and Sorensen]{wingate2022prompt}
David Wingate, Mohammad Shoeybi, and Taylor Sorensen.
\newblock Prompt compression and contrastive conditioning for controllability and toxicity reduction in language models.
\newblock \emph{arXiv preprint arXiv:2210.03162}, 2022.

\bibitem[Wolf et~al.(2020)Wolf, Debut, Sanh, Chaumond, Delangue, Moi, Cistac, Rault, Louf, Funtowicz, et~al.]{wolf2020transformers}
Thomas Wolf, Lysandre Debut, Victor Sanh, Julien Chaumond, Clement Delangue, Anthony Moi, Pierric Cistac, Tim Rault, Remi Louf, Morgan Funtowicz, et~al.
\newblock Transformers: State-of-the-art natural language processing.
\newblock In \emph{Proceedings of the 2020 conference on empirical methods in natural language processing: system demonstrations}, pp.\  38--45, 2020.

\bibitem[Wu et~al.(2025{\natexlab{a}})Wu, Teng, and Tu]{wu2025parallel}
Haoyi Wu, Zhihao Teng, and Kewei Tu.
\newblock Parallel continuous chain-of-thought with jacobi iteration, 2025{\natexlab{a}}.

\bibitem[Wu et~al.(2025{\natexlab{b}})Wu, Lu, Ren, Hu, Wu, Dai, and Wu]{liang2025singlethread}
Junhong Wu, Jinliang Lu, Zixuan Ren, Gangqiang Hu, Zhi Wu, Dai Dai, and Hua Wu.
\newblock Llms are single-threaded reasoners: Demystifying the working mechanism of soft thinking.
\newblock \emph{arXiv preprint arXiv:2508.03440}, 2025{\natexlab{b}}.

\bibitem[Xiao et~al.(2023)Xiao, Tian, Chen, Han, and Lewis]{xiao2023efficient}
Guangxuan Xiao, Yuandong Tian, Beidi Chen, Song Han, and Mike Lewis.
\newblock Efficient streaming language models with attention sinks.
\newblock \emph{arXiv preprint arXiv:2309.17453}, 2023.

\bibitem[Xu et~al.(2025{\natexlab{a}})Xu, Guo, Zeng, and Miao]{xu2025softcot}
Yige Xu, Xu~Guo, Zhiwei Zeng, and Chunyan Miao.
\newblock Softcot: Soft chain-of-thought for efficient reasoning with llms.
\newblock \emph{arXiv preprint arXiv:2502.12134}, 2025{\natexlab{a}}.

\bibitem[Xu et~al.(2025{\natexlab{b}})Xu, Guo, Zeng, and Miao]{xu2025softcot++}
Yige Xu, Xu~Guo, Zhiwei Zeng, and Chunyan Miao.
\newblock Softcot++: Test-time scaling with soft chain-of-thought reasoning.
\newblock \emph{arXiv preprint arXiv:2505.11484}, 2025{\natexlab{b}}.

\bibitem[Yang et~al.(2025)Yang, Li, Yang, Zhang, Hui, Zheng, Yu, Gao, Huang, Lv, et~al.]{yang2025qwen3}
An~Yang, Anfeng Li, Baosong Yang, Beichen Zhang, Binyuan Hui, Bo~Zheng, Bowen Yu, Chang Gao, Chengen Huang, Chenxu Lv, et~al.
\newblock Qwen3 technical report.
\newblock \emph{arXiv preprint arXiv:2505.09388}, 2025.

\bibitem[Yao et~al.(2023{\natexlab{a}})Yao, Yu, Zhao, Shafran, Griffiths, Cao, and Narasimhan]{yao2023tree}
Shunyu Yao, Dian Yu, Jeffrey Zhao, Izhak Shafran, Tom Griffiths, Yuan Cao, and Karthik Narasimhan.
\newblock Tree of thoughts: Deliberate problem solving with large language models.
\newblock \emph{Advances in neural information processing systems}, 36:\penalty0 11809--11822, 2023{\natexlab{a}}.

\bibitem[Yao et~al.(2023{\natexlab{b}})Yao, Zhao, Yu, Du, Shafran, Narasimhan, and Cao]{yao2023react}
Shunyu Yao, Jeffrey Zhao, Dian Yu, Nan Du, Izhak Shafran, Karthik Narasimhan, and Yuan Cao.
\newblock React: Synergizing reasoning and acting in language models.
\newblock In \emph{International Conference on Learning Representations (ICLR)}, 2023{\natexlab{b}}.

\bibitem[Yentinglin(2025)]{hf_aime25_dataset}
Yentinglin.
\newblock Aime 2025 (american invitational mathematics examination 2025).
\newblock Hugging Face dataset, 2025.
\newblock URL \url{https://huggingface.co/datasets/yentinglin/aime_2025}.

\bibitem[Yuan et~al.(2025{\natexlab{a}})Yuan, Chen, Yu, Shi, Jin, Lee, and Mitra]{yuan2025mitigating}
Xiangchi Yuan, Xiang Chen, Tong Yu, Dachuan Shi, Can Jin, Wenke Lee, and Saayan Mitra.
\newblock Mitigating forgetting between supervised and reinforcement learning yields stronger reasoners.
\newblock \emph{arXiv preprint arXiv:2510.04454}, 2025{\natexlab{a}}.

\bibitem[Yuan et~al.(2025{\natexlab{b}})Yuan, Zhang, Liu, Shi, Vosoughi, and Lee]{yuan2025superficial}
Xiangchi Yuan, Chunhui Zhang, Zheyuan Liu, Dachuan Shi, Soroush Vosoughi, and Wenke Lee.
\newblock Superficial self-improved reasoners benefit from model merging.
\newblock \emph{arXiv preprint arXiv:2503.02103}, 2025{\natexlab{b}}.

\bibitem[Yue et~al.(2025)Yue, Jin, Zeng, Zhuang, Qin, Yoon, Shang, Han, and Wang]{yue2025hybrid}
Zhenrui Yue, Bowen Jin, Huimin Zeng, Honglei Zhuang, Zhen Qin, Jinsung Yoon, Lanyu Shang, Jiawei Han, and Dong Wang.
\newblock Hybrid latent reasoning via reinforcement learning.
\newblock \emph{arXiv preprint arXiv:2505.18454}, 2025.

\bibitem[Zeng et~al.(2025)Zeng, Song, Huang, Wang, Li, He, Wang, Li, and Lin]{zeng2025pretraining}
Boyi Zeng, Shixiang Song, Siyuan Huang, Yixuan Wang, He~Li, Ziwei He, Xinbing Wang, Zhiyu Li, and Zhouhan Lin.
\newblock Pretraining language models to ponder in continuous space, 2025.

\bibitem[Zhang et~al.(2025)Zhang, He, Yan, Shen, Zhao, Wang, Shen, and Wang]{zhang2025softthinking}
Zhen Zhang, Xuehai He, Weixiang Yan, Ao~Shen, Chenyang Zhao, Shuohang Wang, Yelong Shen, and Xin~Eric Wang.
\newblock Soft thinking: Unlocking the reasoning potential of llms in continuous concept space.
\newblock \emph{arXiv preprint arXiv:2505.15778}, 2025.

\bibitem[Zhou et~al.(2022)Zhou, Sch{\"a}rli, Hou, Wei, Scales, Wang, Schuurmans, Cui, Bousquet, Le, et~al.]{zhou2022least}
Denny Zhou, Nathanael Sch{\"a}rli, Le~Hou, Jason Wei, Nathan Scales, Xuezhi Wang, Dale Schuurmans, Claire Cui, Olivier Bousquet, Quoc Le, et~al.
\newblock Least-to-most prompting enables complex reasoning in large language models.
\newblock \emph{arXiv preprint arXiv:2205.10625}, 2022.

\bibitem[Zhu et~al.(2025{\natexlab{a}})Zhu, Hao, Hu, Jiao, Russell, and Tian]{zhu2025reasoning}
Hanlin Zhu, Shibo Hao, Zhiting Hu, Jiantao Jiao, Stuart Russell, and Yuandong Tian.
\newblock Reasoning by superposition: A theoretical perspective on chain of continuous thought.
\newblock \emph{arXiv preprint arXiv:2505.12514}, 2025{\natexlab{a}}.

\bibitem[Zhu et~al.(2025{\natexlab{b}})Zhu, Peng, Cheng, Qu, Huang, Zhu, Wang, Xue, Zhang, Shan, et~al.]{zhu2025survey}
Rui-Jie Zhu, Tianhao Peng, Tianhao Cheng, Xingwei Qu, Jinfa Huang, Dawei Zhu, Hao Wang, Kaiwen Xue, Xuanliang Zhang, Yong Shan, et~al.
\newblock A survey on latent reasoning.
\newblock \emph{arXiv preprint arXiv:2507.06203}, 2025{\natexlab{b}}.

\end{thebibliography}
\bibliographystyle{template/iclr2026_conference}

\appendix
\newpage

\section{Use of LLMs Disclosure}

We employed GPT-5~\citep{openai2025gpt5systemcard} from OpenAI to assist with language polishing in order to improve the readability of the paper. We affirm that large language models were not misused intentionally in any part of this work. All intellectual contributions are attributed to the human authors, and the results presented in this paper are entirely the product of human research efforts.

\section{Supplementary Details}

\subsection{SwiReasoning Implementation}
\label{supp_imp}

\begin{algorithm}[h]
\caption{\textproc{SwiReasoning}}
\label{alg:swi}
\begin{algorithmic}[1]
\Require Question $x_{1:n}$, model $\mathcal{M}$, max steps $T_{\max}$, coefficient $\alpha_0$, coefficient $\beta_0$, dwell window $W_{\mathrm{E}\to\mathrm{L}}$, 
{\color{steelBlue}max switches $C_{\max}$, and answer budget $B$}
\Ensure Answer $y_{1:m}$

\State \textbf{Init:} Mode $m_0 \gets \text{Latent}$, {\color{steelBlue}switch counter $C \gets 0$, injection queue $Q \gets \emptyset$, budget flag $b \gets -1$}

\For{$t=1$ to $T_{\max}$}
  \State $\ell_t \gets \mathcal{M}(x_{1:t-1})$;\quad $p_t \gets \mathrm{softmax}(\ell_t)$;\quad $H_t \gets -\sum_v p_t[v]\log p_t[v]$ \Comment{Entropy}

  \color{steelBlue}
  \If{$Q \neq \emptyset$} \Comment{Token injection (convergence/termination prefix)}
    \State $x_t \gets Q.\mathrm{pop}()$
    \If{$b = 0$} \State \textbf{break} \EndIf
    \If{$b > 0$} \State $b \gets b-1$ \EndIf
    \State \textbf{continue}
  \EndIf
  \color{black}

  {\If{$t=1$} 
     \State $\bar H \gets H_1$; \quad $\Delta t \gets 0$
  \EndIf
  \If{$m_{t-1}=\text{Latent}$ \textbf{and} $H_t < \bar H$} \Comment{Mode switching (Sec.\,3.3)}
     \State $m_t \gets \text{Explicit}$; \quad $\bar H \gets H_t$; \quad $\Delta t \gets 0$; \quad {\color{steelBlue}$C \gets C+1$} 
  \ElsIf{$m_{t-1}=\text{Explicit}$ \textbf{and} $H_t > \bar H$ \textbf{and} $\Delta t \ge W_{\mathrm{E}\to\mathrm{L}}$}
     \State $m_t \gets \text{Latent}$; \quad $\bar H \gets H_t$; \quad $\Delta t \gets 0$
  \Else
     \State $m_t\leftarrow m_{t-1}$;\quad $\Delta t\gets \Delta t+1$
  \EndIf}

  \color{steelBlue}
  \If{$m_t=\text{Explicit}$ \textbf{and} $\frac{1}{2}C_{\max} \leq C \leq C_{\max}$} \Comment{Switch count control (Sec.\,3.4)}
     \State $Q \gets [\,\ID{\tokEnd}\,]$ \Comment{Convergence trigger}
  \ElsIf{$m_t=\text{Explicit}$ \textbf{and} $C > C_{\max}$}
     \State $Q \gets [\,\ID{\text{``}\tokEnd\text{\textbackslash n\textbackslash n The final answer is''}}\,]$;\quad $b \gets B$
     \Comment{Termination trigger}
  \EndIf
  \color{black}

  \If{$m_t=\text{Explicit}$ \textbf{and} $\Delta t > 0$} 
     \State $x_t \gets \arg\max_v p_t[v]$ or Sampling
  \Else 
     \State $\tilde e_t \gets \sum_v p_t[v]\,E[v]$ 
     \If{$m_t=\text{Latent}$ \textbf{and} $\Delta t = 0$} \Comment{Thinking-related signal mixing}
        \State $\alpha_t=\alpha_0+(1-\alpha_0)\frac{t}{T_{\max}}$
        \State $\tilde e_t \gets \alpha_t\,\tilde e_t + (1-\alpha_t)\,e_{\tokThink}$
     \EndIf
     \If{$m_t=\text{Explicit}$ \textbf{and} $\Delta t = 0$} \Comment{Thinking-related signal mixing}
        \State $\beta_t=\beta_0+(1-\beta_0)\frac{t}{T_{\max}}$
        \State $\tilde e_t \gets \beta_t\,\tilde e_t + (1-\beta_t)\,e_{\tokEnd}$
     \EndIf
     \State $x_t \gets \tilde e_t$ \Comment{Soft embeddings feed as inputs}
  \EndIf

  \If{$x_t=\text{\textless EOS\textgreater}$} \State \textbf{break} \EndIf
\EndFor

\State Extract answer $y$ from $x_{n+1:t}$
\State \Return $y$
\end{algorithmic}
\end{algorithm}

Alg.~\ref{alg:swi} provides a detailed implementation of \swireasoning, where the implementation for mode switching is written in black and switch count control for token efficiency is outlined in \textcolor{steelBlue}{blue}.

\subsection{Benchmark Details}
\label{supp_benchmark}


We conduct evaluations on 11 reasoning benchmarks, including GSM8K~\citep{cobbe2021training}, Math500~\citep{hendrycks2021measuring}, AIME 2024~\citep{hf_aime24_dataset}, AIME 2025~\citep{hf_aime25_dataset} for mathematical reasoning; GPQA Diamond~\citep{rein2024gpqa} for STEM reasoning; HumanEval~\citep{chen2021evaluating}, LeetCode-Contest~\citep{guo2024deepseekcoder}, MBPP~\citep{austin2021program}, LiveCodeBench~\citep{jain2024livecodebench} for coding reasoning; and 2WikiMultihopQA~\citep{ho2020constructing}, CommonsenseQA~\citep{talmor2019commonsenseqa} for general reasoning.

\begin{itemize}
    \item \textbf{GSM8K}: We use the test set of 1,319 grade-school math word problems, designed to evaluate multi-step arithmetic reasoning in natural language. \hf: \url{https://huggingface.co/datasets/openai/gsm8k}.
    \item \textbf{MATH500}: A curated set of 500 problems from the MATH dataset, covering diverse high-school competition-level mathematics domains such as algebra, geometry, and number theory. \hf: \url{https://huggingface.co/datasets/HuggingFaceH4/MATH-500}.
    \item \textbf{AIME 2024}: Contains 30 problems from the 2024 American Invitational Mathematics Examination, each requiring a concise numeric answer and reflecting competition-level difficulty. \hf: \url{https://huggingface.co/datasets/HuggingFaceH4/aime_2024}.
    \item \textbf{AIME 2025}: Contains 30 problems from the 2025 American Invitational Mathematics Examination, continuing the focus on competition-style math reasoning with challenging questions that test symbolic and logical skills. \hf: \url{https://huggingface.co/datasets/yentinglin/aime_2025}.
    \item \textbf{GPQA Diamond}: A high-quality subset of about 198 carefully verified questions, focusing on STEM disciplines including mathematics, physics, chemistry, biology, and computer science, designed to evaluate expert-level factual knowledge and reasoning ability. \hf: \url{https://huggingface.co/datasets/hendrydong/gpqa_diamond_mc}.
    \item \textbf{HumanEval}: A set of 164 hand-written Python programming problems. Each problem provides a function signature and a short docstring specification, and models are evaluated by executing generated code against unit tests. \hf: \url{https://huggingface.co/datasets/openai/openai_humaneval}.
    \item \textbf{LeetCode-Contest}: A benchmark for evaluating code LLMs, consisting of 180 algorithm problems of different difficulties from LeetCode contests. The evaluation is based on whether the generated solution passes the provided tests. \hf: \url{https://huggingface.co/datasets/TechxGenus/LeetCode-Contest}.
    \item \textbf{MBPP}: We use the test set of 257 Python programming problems. Each problem includes a natural language prompt, a reference solution, and unit tests. The model is scored by running generated code on the tests. \hf: \url{https://huggingface.co/datasets/google-research-datasets/mbpp}.
    \item \textbf{LiveCodeBench}: The dataset contains 279 problems, and the prompt instructs the model to generate a correct Python program that matches the specification and passes all tests. \hf: \url{https://huggingface.co/datasets/PrimeIntellect/LiveCodeBench-v5}.
    \item \textbf{2WikiMultihopQA}: We use the 2WikiMultihopQA set with 200 problems from LongBench~\citep{bai2024longbench}. The problems require combining evidence from multiple Wikipedia documents to reach the final answer. It tests cross-document reasoning rather than single passage lookup. \hf: \url{https://huggingface.co/datasets/zai-org/LongBench}.
    \item \textbf{CommonsenseQA}: We use the test set of 1,221 multiple-choice QA questions that require commonsense knowledge. Questions are constructed from ConceptNet relations and each question comes with one correct answer and several distractor options. \hf: \url{https://huggingface.co/datasets/tau/commonsense_qa}.
\end{itemize}

To provide LLMs with sufficient thinking space, following the same settings as Qwen3~\citep{yang2025qwen3}, we set the maximum generation length to 32,768 tokens for GSM8K, Math500, GPQA Diamond, HumanEval, LeetCode-Contest, MBPP, LiveCodeBench, 2WikiMultihopQA, and CommonsenseQA benchmarks, and 38,912 tokens for AIME 2024 and AIME 2025 benchmarks.

We repeat the experiments eight times and report the average accuracy for both \swireasoning and other baselines on the AIME 2024 and AIME 2025 benchmarks.

\subsection{Best Practice for Hyperparameters}
\label{supp_hyperparameter}


\begin{table*}[h]
\small
\centering
\captionsetup{font={small}}
\setlength{\tabcolsep}{2pt}
\caption{Hyperparameters for mode switching across datasets and models. $W$ and $\beta_0$ are fixed across all scenarios, while $\alpha_0$ provides users with flexibility for adjustment depending on the task.}
\label{tab:hp_per_dataset}
\begin{tabular}{llcccc}
\toprule
\multirow{2}{*}[-1pt]{\textbf{Hyperparameter}} & \multirow{2}{*}[-1.7pt]{\textbf{Dataset}} & \multicolumn{4}{c}{\textbf{Model}} \\
\cmidrule(lr){3-6}
 & & \textit{Qwen3-1.7B} & \textit{Qwen3-8B} & \textit{Qwen3-32B} & \textit{DeepSeek-R1-Distill-Llama-8B} \\
\cmidrule(r){1-2} \cmidrule(r){3-6}
\multirow{11}{*}{$W$ (window size)}
 & GSM8K        & \multicolumn{4}{c}{\multirow{11}{*}{512 \text{(fixed for all)}}} \\
 & MATH500      & \multicolumn{4}{c}{} \\
 & AIME2024     & \multicolumn{4}{c}{} \\
 & AIME2025     & \multicolumn{4}{c}{} \\
 & GPQA Diamond & \multicolumn{4}{c}{} \\
 & HumanEval           & \multicolumn{4}{c}{} \\
 & LeetCode-Contest    & \multicolumn{4}{c}{} \\
 & MBPP                & \multicolumn{4}{c}{} \\
 & LiveCodeBench       & \multicolumn{4}{c}{} \\
 & 2WikiMultihopQA     & \multicolumn{4}{c}{} \\
 & CommonsenseQA       & \multicolumn{4}{c}{} \\
 \cmidrule(r){1-2} \cmidrule(r){3-6}
\multirow{11}{*}{$\alpha_{0}$ (user-exposed)}
 & GSM8K        & 0.6 & 0.5 & 0.5 & 0.1 \\
 & MATH500      & 0.5 & 1.0 & 0.5 & 0.5 \\
 & AIME2024     & 0.5 & 0.9 & 1.0 & 0.65 \\
 & AIME2025     & 0.3 & 0.9 & 0.9 & 0.7 \\
 & GPQA Diamond & 1.0 & 1.0 & 0.6 & 0.7 \\
 & HumanEval           & \multirow{6}{*}{/} & 0.5 & \multirow{6}{*}{/} & \multirow{6}{*}{/} \\
 & LeetCode-Contest    & & 1.0 & & \\
 & MBPP                & & 0.9 & & \\
 & LiveCodeBench       & & 0.9 & & \\
 & 2WikiMultihopQA     & & 0.7 & & \\
 & CommonsenseQA       & & 0.9 & & \\
\cmidrule(r){1-2} \cmidrule(r){3-6}
\multirow{11}{*}{$\beta_{0}$}
 & GSM8K        & \multicolumn{4}{c}{\multirow{11}{*}{0.7 \text{(fixed for all)}}} \\
 & MATH500      & \multicolumn{4}{c}{} \\
 & AIME2024     & \multicolumn{4}{c}{} \\
 & AIME2025     & \multicolumn{4}{c}{} \\
 & GPQA Diamond & \multicolumn{4}{c}{} \\
& HumanEval           & \multicolumn{4}{c}{} \\
 & LeetCode-Contest    & \multicolumn{4}{c}{} \\
 & MBPP                & \multicolumn{4}{c}{} \\
 & LiveCodeBench       & \multicolumn{4}{c}{} \\
 & 2WikiMultihopQA     & \multicolumn{4}{c}{} \\
 & CommonsenseQA       & \multicolumn{4}{c}{} \\
\bottomrule
\end{tabular}
\end{table*}

In addition to Tab.~\ref{tab:hp_per_dataset}, \swireasoning operates as a straightforward and instant substitution for the \texttt{model.generate()} interface of Huggingface's \textit{transformers}~\citep{wolf2020transformers} package. There are no model parameters or architecture changes, and no training was used in the experiments. For sampling-related hyperparameters and prompt templates, we use the ones recommended by Qwen3 and DeepSeek-R1's technical report~\citep{yang2025qwen3, guo2025deepseek} without modification.

\subsection{Broader Related Work}

\paragraph{Efficient LLM Reasoning.} In terms of improving reasoning efficiency, there are broader techniques including but not limited to KV cache compression~\citep{han2023lm, xiao2023efficient, cai2024pyramidkv, shilacache}, prompt token compression~\citep{wingate2022prompt, jiang2023llmlingua, pan2024llmlingua, shi2023crossget}, speculative decoding~\citep{leviathan2023fast, kim2023speculative, liu2023online, sun2023spectr, chen2024sequoia}, traditional methods such as quantization, pruning, distillation~\citep{shi2023upop, lin2024awq, fu2024amoeballm, yuan2025superficial, yuan2025mitigating}, and system-level optimizations such as FlashAttention~\citep{dao2022flashattention, dao2023flashattention, shah2024flashattention}. Learning-based generation methods, such as COCONUT~\citep{hao2024training} based on supervised fine-tuning, and Long$\otimes$Short~\citep{ning2025not} based on reinforcement learning, shorten the reasoning process by involving additional training.

\swireasoning, however, targets a different axis of efficiency and is not aiming to surpass them. Instead, it saves tokens by dynamically alternating between latent steps and explicit steps and limiting the number of block switches. As such, it is plug-and-play during inference and can be layered on top of the aforementioned techniques for multiplicative gains. 

\section{Supplementary Experiments}

\subsection{Quantitative Analyses of exploration in latent and explicit modes}

\begin{figure*}[ht]
    \captionsetup{font={small}}
    \centering
    \includegraphics[width=1.0\linewidth]{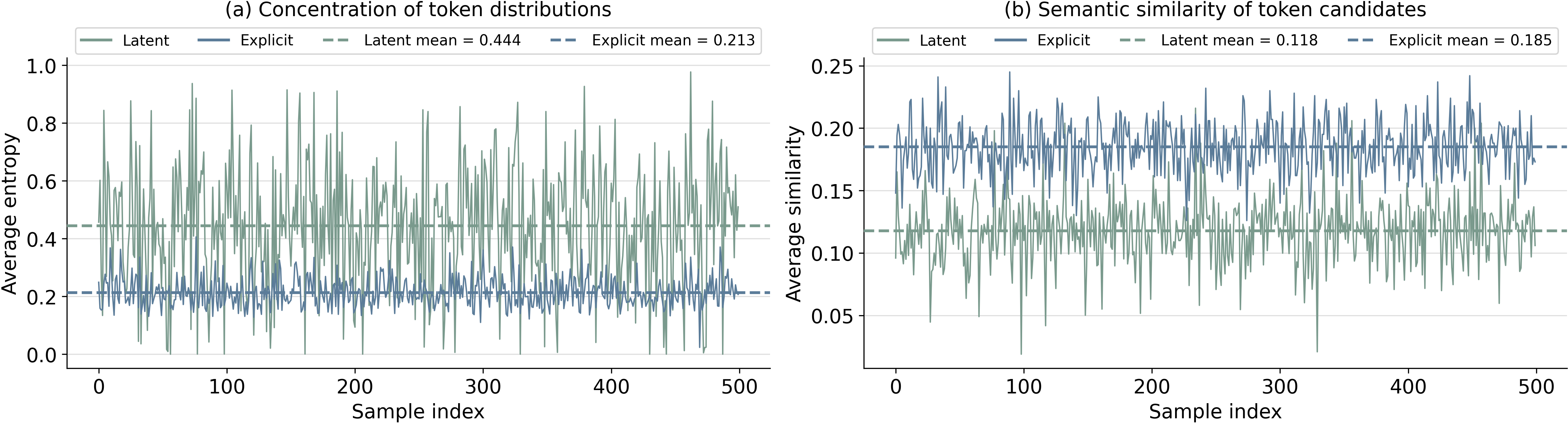}
    \caption{Latent modes enable exploration with less concentrated token distributions and more semantically diverse candidates. In latent modes: (a) the average entropy of the next-token distributions indicates less concentration, and (b) the average pairwise similarity among token candidates indicates more semantic diversity.}
    \label{fig:explore}
\end{figure*}

To quantify exploration in the latent and explicit modes, we analyze two complementary statistics with the Qwen3-8B model on Math500 as visualized in Fig.~\ref{fig:explore}. First, to measure the differences in the concentration level of the next-token distributions between latent and explicit modes, we calculate the average Shannon entropy across steps for each sample. Fig.~(a) shows that the entropy in latent mode is 2.08x higher than in explicit mode on average (0.444 vs. 0.213), indicating that probability mass is spread across more candidate tokens instead of collapsing in latent mode.

Entropy alone does not guarantee semantic diversity, as the semantic distances of different token candidates can be close, even when the probability mass is spread across them. To measure the differences in the semantic similarity among token candidates between latent and explicit modes, we calculate the average pairwise cosine similarity of top-5 token candidates over steps by using their embeddings for each sample. Fig.~(b) shows that the semantic similarity in latent mode is 63.8\% of that in explicit mode on average (0.118 vs. 0.185), showing that token candidates considered under latent modes are more semantically diverse. Taken together, these results quantitatively substantiate that latent modes enable exploration with less concentrated token distributions and more semantically diverse candidates.

\subsection{Efficiency measured in terms of latency and TFLOPs}

\begin{figure*}[ht]
    \captionsetup{font={small}}
    \centering
    \includegraphics[width=1.0\linewidth]{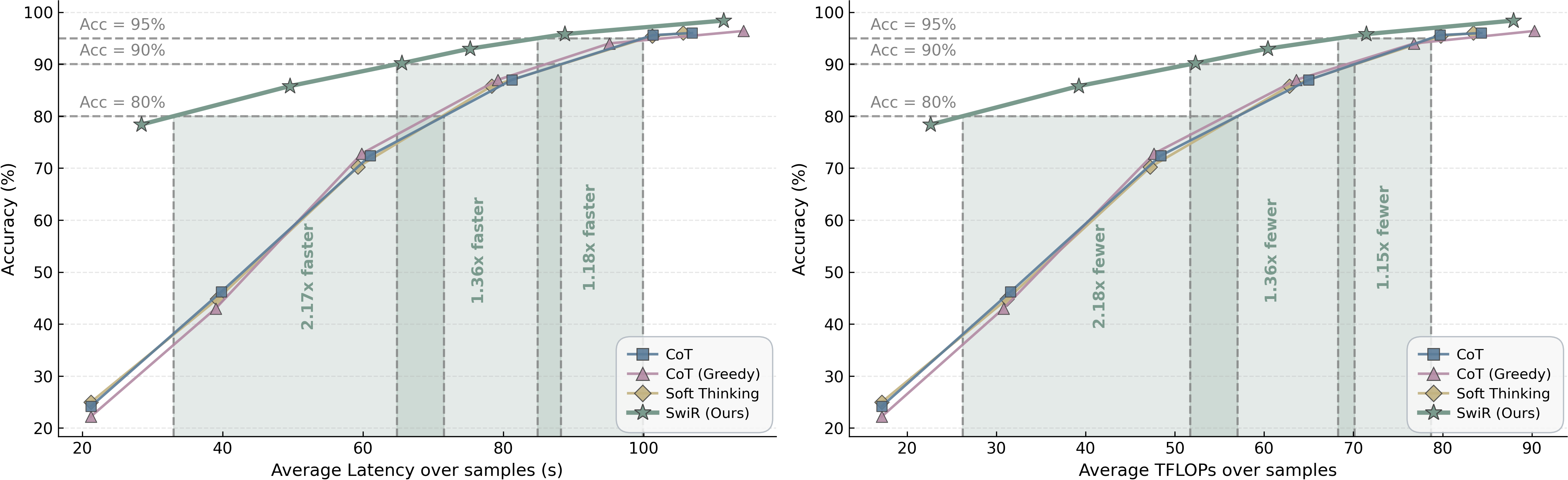}
    \caption{Left: Accuracy vs. wall-clock latency. Right: Accuracy vs. TFLOPs. Dashed horizontal lines mark equal-accuracy targets ($80\%$, $90\%$, $95\%$). Lower values on the x-axes are better.}
    \label{fig:latency_and_tflops}
\end{figure*}

We evaluate efficiency in terms of average wall-clock latency and TFLOPs over samples on MATH500 with Qwen3-8B, comparing CoT, CoT (Greedy), Soft Thinking, and \swireasoning. Fig. \ref{fig:latency_and_tflops} demonstrates that \swireasoning consistently reaches a given accuracy with lower latency and fewer TFLOPs, and the advantage grows as the budget becomes tighter. Specifically, at $90\%$ pass@1 accuracy, \swireasoning uses 1.36× fewer TFLOPs and is 1.36× faster in wall-clock time than CoT. At $80\%$ accuracy, it uses 2.18× fewer TFLOPs and is 2.17× faster. These results confirm that \swireasoning improves efficiency at fixed accuracies, especially under small budgets.

\subsection{Statistics of Switch Counts across Problem Difficulties}

\begin{figure*}[ht]
    \captionsetup{font={small}}
    \centering
    \includegraphics[width=0.8\linewidth]{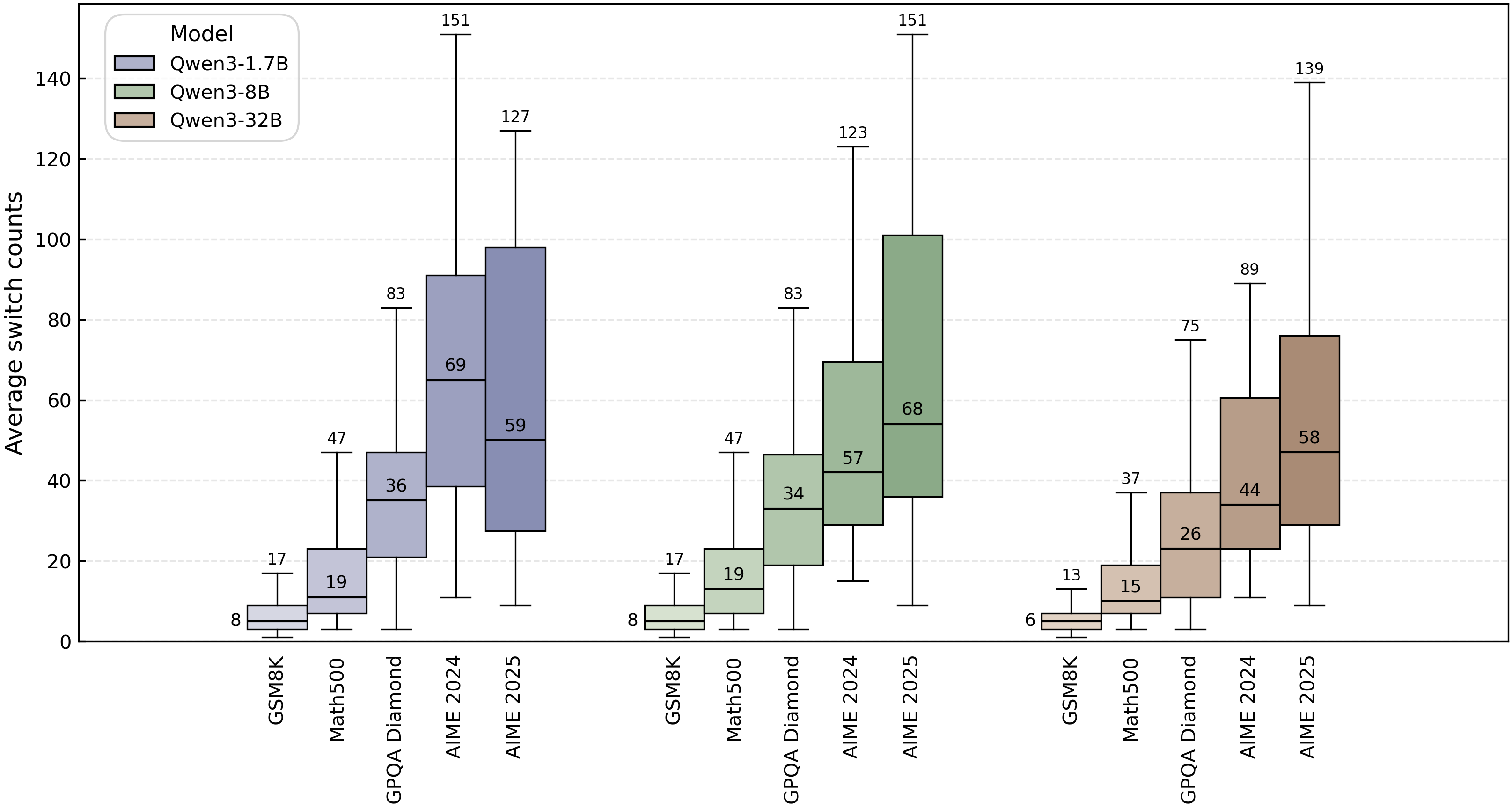}
    \caption{Statistics of switch counts across varying problem difficulties and model scales. The box plots show the distribution of the number of switches for Qwen3-1.7B, Qwen3-8B, and Qwen3-32B on five benchmarks.}
    \label{fig:switch_count}
\end{figure*}

To investigate the switching dynamics, we analyzed the distribution of switch counts across five benchmarks of varying difficulty. Fig.~\ref{fig:switch_count} presents the box plots of switch counts for Qwen3-1.7B, Qwen3-8B, and Qwen3-32B. The switching frequency demonstrates adaptability to problem difficulty. On benchmarks requiring complex reasoning such as AIME 2024 and AIME 2025, the switch counts increase substantially, with medians ranging from 44 to 69, and maximum counts reach 151. The results confirm that SwiReasoning functions as a difficulty-aware reasoning approach.


\subsection{Analysis of Challenges in Specific Scenarios}

\begin{tcolorbox}[examplebox,title={3D Surface Shortest Path}]
\textbf{$\blacktriangleright$Question:} 
A gecko is in a room that is 12 feet long, 10 feet wide and 8 feet tall. The gecko is currently on a side wall ($10^{\prime}$ by $8^{\prime}$), one foot from the ceiling and one foot from the back wall ($12^{\prime}$ by $8^{\prime}$). The gecko spots a fly on the opposite side wall, one foot from the floor and one foot from the front wall. What is the length of the shortest path the gecko can take to reach the fly assuming that it does not jump and can only walk across the ceiling and the walls? Express your answer in simplest radical form.

\medskip

\textbf{$\blacktriangleright$Output:}

To determine the shortest path the gecko can take to reach the fly, we consider the constraints:

\begin{itemize}
  \item The \textbf{gecko} is on a \textbf{side wall} (10' by 8') of a \textbf{12' × 10' × 8'} room.
  \item The \textbf{fly} is on the \textbf{opposite side wall}.
  \item The gecko is \textbf{one foot from the ceiling} and \textbf{one foot from the back wall}.
  \item The fly is \textbf{one foot from the floor} and \textbf{one foot from the front wall}.
  \item The gecko \textbf{can only walk across the ceiling and walls}, not the floor.
\end{itemize}

\noindent\rule{\linewidth}{0.4pt}

\subsubsection*{Step 1: Assign Coordinates}

Let’s define the room in 3D coordinates:

\begin{itemize}
  \item The \textbf{length} (along the x-axis) is 12 feet.
  \item The \textbf{width} (along the y-axis) is 10 feet.
  \item The \textbf{height} (along the z-axis) is 8 feet.
\end{itemize}

Assume:

\begin{itemize}
  \item The \textbf{gecko} is on the \textbf{left wall} (x = 0), one foot from the \textbf{back wall} (y = 1), and one foot from the \textbf{ceiling} (z = 7). So its position is:\\
  \textbf{(x = 0, y = 1, z = 7)}
  \item The \textbf{fly} is on the \textbf{right wall} (x = 12), one foot from the \textbf{front wall} (y = 9), and one foot from the \textbf{floor} (z = 1). So its position is:\\
  \textbf{(x = 12, y = 9, z = 1)}
\end{itemize}

\noindent\rule{\linewidth}{0.4pt}

\subsubsection*{Step 2: Consider the Shortest Path}

The gecko can only walk \textbf{across the ceiling and walls}. The shortest path will involve \textbf{unfolding} the room into a flat 2D plane such that the gecko's path becomes a straight line.

A common and effective approach is to \textbf{unfold} the \textbf{left wall}, \textbf{ceiling}, and \textbf{right wall} into a single plane. This allows the gecko to move from its position on the left wall, across the ceiling, to the fly on the right wall.

\noindent\rule{\linewidth}{0.4pt}

\subsubsection*{Step 3: Transform to a 2D Coordinate System}

In this unfolded net:

\begin{itemize}
  \item The \textbf{left wall} is 10 feet wide (y-axis) and 8 feet tall (z-axis).
  \item The \textbf{ceiling} is 12 feet long (x-axis) and 10 feet wide (y-axis).
  \item The \textbf{right wall} is 10 feet wide (y-axis) and 8 feet tall (z-axis).
\end{itemize}

When we unfold the left wall, ceiling, and right wall into one plane, the gecko's position and the fly's position can be mapped in this 2D plane.

In this 2D representation:

\begin{itemize}
  \item The \textbf{gecko's position} is at (y = 1, z = 7) on the left wall.
  \item The \textbf{fly's position} is at (y = 9, z = 1) on the right wall.
\end{itemize}

However, since the gecko is on the \textbf{left wall}, and the fly is on the \textbf{right wall}, and the path goes \textbf{across the ceiling}, we can consider the \textbf{horizontal distance} as the \textbf{length of the room (12 feet)}, and the \textbf{vertical distance} as the \textbf{difference in heights from the gecko to the ceiling (1 foot)} and from the ceiling to the fly (7 feet), totaling \textbf{8 feet}.

So, the total path forms a \textbf{right triangle} with: \textcolor{revision}{[Incorrect reasoning trace here that does not adhere to a “only walk across the ceiling and the walls" constraint]}

\begin{itemize}
  \item One leg: \textbf{12 feet} (length of the room)
  \item The other leg: \textbf{8 feet} (vertical distance from gecko to ceiling + ceiling to fly) 
\end{itemize}

\noindent\rule{\linewidth}{0.4pt}

\subsubsection*{Step 4: Compute the Shortest Distance}

Using the Pythagorean theorem:

$$
\text{Distance} = \sqrt{12^2 + 8^2} = \sqrt{144 + 64} = \sqrt{208} = \sqrt{16 \cdot 13} = 4\sqrt{13}
$$

\noindent\rule{\linewidth}{0.4pt}

\subsubsection*{Final Answer}

$$
\boxed{4\sqrt{13}}
$$

\medskip

\textbf{$\blacktriangleright$Correct Answer:} $2\sqrt{113}$

\end{tcolorbox}

While \swireasoning demonstrates robust generalization across diverse domains, we identify a specific category of tasks, spatial reasoning with rigid topological constraints, which may be challenging for latent-based methods. A representative case is the “3D Surface Shortest Path” problem from the Math500 benchmark shown above. In this problem, a gecko must find the shortest path on a 3D room’s surface. The reasoning trace reveals that \swireasoning correctly identifies the high-level strategy: unfolding the 3D room into a 2D plane. However, it struggles to maintain the specific topological “only walk across the ceiling and the walls” constraint during latent transitions.

We attribute this behavior to a trade-off inherent in latent reasoning:
\begin{itemize}[leftmargin=*,labelsep=0.5em]
  \item Smoothing vs. Precision: Latent reasoning operates by smoothing probability distributions to enable diverse semantic exploration. While this is beneficial for logical reasoning, it may be detrimental for tasks with precise constraints, such as rigid geometric topology (\textit{e.g.}, defining exactly which wall edge connects to which).
  \item Blurring Constraints: During the latent mode, the strict “only walk across the ceiling and the walls” constraint is blurred due to the probability mass spreading. As a result, when the model switches back to explicit mode, it may reason with an invalid path (\textit{e.g.}, calculating a direct Euclidean distance through prohibited space), leading to an incorrect answer ($4\sqrt{13}$ vs. true $2\sqrt{113}$).
\end{itemize}

Overall, this suggests that while \swireasoning excels at semantic and logical exploration, a few tasks requiring rigid geometric constraint satisfaction, where fuzzy latent exploration may interfere with precise geometric execution, could still be challenging.

\subsection{Ablation Study on the Convergence Triggers w.r.t. Termination Ones}
\label{abl:triggers}

\begin{table}[ht]
\small
\centering
\captionsetup{font={small}}
\setlength{\tabcolsep}{6pt}
\caption{Ablation study on the relative position of convergence and termination triggers. The convergence trigger is set at different fractions of the termination count $T$ (where $T=C_{\max}$). The results demonstrate that the default $\frac{1}{2}T$ offers a balanced choice, avoiding the significant accuracy loss of earlier triggers while being more efficient than later ones.}
\label{tab:triggers}
\begin{tabular}{lccccc}
\toprule
\textbf{Convergence w.r.t. Termination} & $C = \tfrac{1}{4}T$ & $C = \tfrac{1}{3}T$ & $C = \tfrac{1}{2}T$ & $C = \tfrac{2}{3}T$ & $C = \tfrac{3}{4}T$ \\
\cmidrule{1-1} \cmidrule(l){2-6} 
\multicolumn{6}{c}{$C_{\max}=8$} \\
\cmidrule{1-1} \cmidrule(l){2-6} 
Accuracy (\%)        & 76.60 & 79.20 & 81.00 & 84.20 & 84.20 \\
Generation Length    & 1428  & 1817  & 2222  & 2743  & 2743  \\
\cmidrule{1-1} \cmidrule(l){2-6} 
\multicolumn{6}{c}{$C_{\max}=20$} \\
\cmidrule{1-1} \cmidrule(l){2-6} 
Accuracy (\%)        & 84.00 & 85.80 & 87.80 & 88.20 & 88.80 \\
Generation Length    & 2788  & 3278  & 3597  & 4002  & 3943  \\
\bottomrule
\end{tabular}
\end{table}

To analyze the positions of convergence triggers w.r.t. termination ones, we conducted an ablation study using Qwen3-1.7B on the Math500 benchmark. We fixed the termination trigger at two representative budgets ($C_{\max}=8$ and $C_{\max}=20$) and varied the position of the convergence trigger relative to the termination step. Specifically, if $T$ denotes the switch count at termination (\textit{i.e.}, $T=C_{\max}$), we tested setting the convergence trigger at $C \in \{\frac{1}{4}T, \frac{1}{3}T, \frac{1}{2}T, \frac{2}{3}T, \frac{3}{4}T\}$. Tab.~\ref{tab:triggers} presents a clear trade-off between accuracy and token efficiency:
\begin{itemize}[leftmargin=*,labelsep=0.5em]
  \item Early convergence ($\frac{1}{4}T, \frac{1}{3}T$): triggering convergence too early significantly reduces generation length but leads to a notable drop in accuracy (\textit{e.g.}, at $C_{\max}=20$, accuracy drops from $87.80\%$ to $84.00\%$ when moving from $\frac{1}{2}T$ to $\frac{1}{4}T$). This suggests that forcing the model to converge before sufficient exploration prevents it from finding correct solutions for harder problems.
  \item Late convergence ($\frac{2}{3}T, \frac{3}{4}T$): delaying the convergence trigger yields marginal accuracy gains (\textit{e.g.}, +$1.0\%$ from $\frac{1}{2}T$ to $\frac{3}{4}T$ at $C_{\max}=20$) but comes at the cost of increased token consumption.
  \item Balanced default ($\frac{1}{2}T$): the default setting of $\frac{1}{2}C_{\max}$ effectively strikes a balance. It captures the majority of the accuracy while maintaining reasonable token efficiency. Overall, $C=\frac{1}{2}T$ serves as a sweet spot for general-purpose reasoning under budget constraints.
\end{itemize}

\subsection{Ablation Study on Isolating the Effect of Latent under Small Budgets}

\begin{table*}[ht]
\small
\centering
\captionsetup{font={small}}
\setlength{\tabcolsep}{0.42mm}
\caption{Ablation study on isolating the effect of latent under small token budgets. "SwiR w/o latent" retains the switch count controller but disables latent mode.}
\label{tab:no_latent}

\begin{tabular}{l r l  r l  r l  r l  r l  r l}
\toprule
\multirow{3}{*}[2pt]{\makecell{\textbf{Method}}} &
\multicolumn{2}{c}{\makecell{GSM8K}} &
\multicolumn{2}{c}{\makecell{MATH\\500}} &
\multicolumn{2}{c}{\makecell{GPQA\\Diamond}} &
\multicolumn{2}{c}{\makecell{AIME\\2024}} &
\multicolumn{2}{c}{\makecell{AIME\\2025}} &
\multicolumn{2}{c}{\makecell{\textbf{Average}}} \\
\cmidrule{2-11}\cmidrule(l){12-13}

\multicolumn{13}{c}{\textit{Accuracy (\%)}} \\
\midrule
SwiR w/o latent   & 85.44 &
                  & 68.60 &
                  & 26.26 &
                  &  2.50 &
                  &  0.42 &
                  & 36.64 & \multicolumn{1}{c}{\scriptsize\textcolor{gray}{$+0.00$}} \\
\rowcolor{gray!20}\textit{\textbf{SwiR}}
                  & \textbf{86.80} & \Deltabar{ 1.36}
                  & \textbf{72.40} & \Deltabar{ 3.80}
                  & \textbf{29.80} & \Deltabar{ 3.54}
                  & \textbf{ 5.83} & \Deltabar{ 3.33}
                  & \textbf{ 4.58} & \Deltabar{ 4.16}
                  & \textbf{39.88} & \Deltabar{ 3.24} \\
\midrule

\multicolumn{13}{c}{\textit{Generation Length}} \\
\midrule
SwiR w/o latent   &  \textbf{816}  &
                  & 1207  &
                  & 1032  &
                  &  835  &
                  &  791  &
                  &  936  & \multicolumn{1}{c}{\scriptsize\textcolor{gray}{$+0.00$}} \\
\textit{\textbf{SwiR}}
                  & \textbf{816}  &
                  & \textbf{1189} &
                  & \textbf{ 933} &
                  & \textbf{ 812} &
                  & \textbf{ 787} &
                  & \textbf{ 907} & \multicolumn{1}{c}{\scriptsize\textcolor{oliveGreen}{$-3.10\%$}} \\
\bottomrule
\end{tabular}
\end{table*}

To further evaluate the effect of latent modes at small budgets, we used Qwen3-1.7B and compared the full \swireasoning against a variant where latent reasoning is disabled (SwiR w/o latent). In the "w/o latent" setting, the model is forced to remain in explicit mode, but retains the same switch count controller (functioning as a step-limit controller for inserting the proposed convergence and termination triggers). We evaluated both methods under small token budgets. 

Tab.~\ref{tab:no_latent} shows that disabling latent leads to a consistent degradation in performance across all benchmarks. Under lower or the same token usage, the full SwiReasoning achieves an average accuracy improvement of $+3.24\%$ over the ablation. On challenging benchmarks like AIME2025, the contribution of latent is particularly pronounced ($+4.16\%$). These results show that the latent itself provides gains that cannot be attributed to token controller mechanisms alone.

\subsection{Supplementary Component-wise Ablation Study}
\label{abl:component}

\begin{table*}[ht]
\small
\centering
\captionsetup{font={small}}
\setlength{\tabcolsep}{4pt}
\caption{Supplementary ablation studies on switch window mechanisms (including window size, symmetry vs. asymmetry) and signal mixing.}
\label{tab:window_and_mixing}

\begin{tabular}{lccc}
\toprule
\textbf{Method} & GSM8K & Math500 & Average \\
\midrule
\textit{SwiR} w/ window size 0 & 89.00\% & 89.20\% & 89.10\% \\
\cmidrule(r){1-1}\cmidrule{2-4}
\textit{SwiR} w/ symmetry window size 2 & 89.31\% & 90.00\% & 89.66\% \\
\textit{SwiR} w/ symmetry window size 4 & \textbf{90.83\%} & 89.20\% & 90.02\% \\
\textit{SwiR} w/ symmetry window size 8 & 90.60\% & 91.20\% & 90.90\% \\
\textit{SwiR} w/ symmetry window size 16 & 88.93\% & 92.00\% & 90.47\% \\
\textit{SwiR} w/ symmetry window size 32 & 90.14\% & 91.00\% & 90.57\% \\
\cmidrule(r){1-1}\cmidrule{2-4}
\textit{SwiR} w/ asymmetry window size 64 & 89.69\% & 92.60\% & 91.15\% \\
\textit{SwiR} w/ asymmetry window size 128 & 90.45\% & 91.00\% & 90.72\% \\
\textit{SwiR} w/ asymmetry window size 256 & 89.76\% & 90.80\% & 90.28\% \\
\rowcolor{gray!20} \textcolor{oliveGreen}{\textbf{\textit{SwiR} w/ asymmetry window size = 512}} \textcolor{oliveGreen}{\cmark}  & \textbf{90.83\%} & \textbf{93.00\%} & \textbf{91.92\%} \\
\textit{SwiR} w/ asymmetry window size = 1024 & \textbf{90.83\%} & 91.20\% & 91.01\% \\
\cmidrule(r){1-1}\cmidrule{2-4}
\textit{SwiR} w/o mixing & 90.44\% & 92.20\% & 91.32\% \\
\bottomrule
\end{tabular}
\end{table*}

To quantify the marginal contribution of each proposed mechanism, we conducted a component-wise ablation study using Qwen3-1.7B on GSM8K and MATH500 benchmarks. We compared the default asymmetric window ($W_{E \rightarrow L}=512, W_{L \rightarrow E}=0$) against variations with no window ($W=0$) and symmetric windows of varying sizes. Results in Tab.~\ref{tab:window_and_mixing} reveal that 
\begin{itemize}[leftmargin=*,labelsep=0.5em]
  \item Necessity of a window: Removing the window constraint completely leads to a significant performance drop (Avg: $89.10\%$ vs. $91.92\%$), due to rapid oscillations between modes preventing coherent reasoning chains from forming.
  \item Benefit of asymmetry: The default asymmetric setting outperforms all symmetric configurations. While symmetric windows (\textit{e.g.}, size 8) improve over no window, they fail to match the peak performance of the asymmetric design. This validates our hypothesis that explicit reasoning requires a consolidation period ($W_{E \rightarrow L}$), while latent reasoning benefits from immediate exit upon confidence recovery ($W_{L \rightarrow E}=0$) to reduce the risks of introducing spurious signals that may mislead the model.
\end{itemize}

Tab.~\ref{tab:window_and_mixing} also indicates that disabling the signal mixing results in a consistent accuracy decrease (Avg: $91.32\%$ vs. $91.92\%$). This suggests that the signal mixing helps the model better orient its internal state during mode transitions and contributes positively to the final accuracy.

\begin{table*}[ht]
\small
\centering
\captionsetup{font={small}}
\setlength{\tabcolsep}{4pt}
\caption{Supplementary ablation study on the maximum switch count $C_{\max}$.  We evaluate the trade-off between accuracy and generation length by varying $C_{max}$ from 4 to unlimited.}
\label{tab:cmax}

\begin{tabular}{lcccccc}
\toprule
$\mathbf{C_{\max}}$ & 4 & 8 & 12 & 16 & 20 & 24 \\
\cmidrule(r){1-1}\cmidrule{2-7}
Accuracy           & 72.40\% & 81.00\% & 85.00\% & 86.80\% & 87.80\% & 87.80\% \\
Generation Length  & 1189    & 2222    & 2914    & 3280    & 3597    & 3862    \\
\midrule
$\mathbf{C_{\max}}$ & 28 & 32 & 36 & 40 & 44 & $\infty$ \\
\cmidrule(r){1-1}\cmidrule{2-7}
Accuracy           & 89.00\% & 89.20\% & 89.80\% & 90.60\% & 91.80\% & \textbf{93.00\%} \\
Generation Length  & 4068    & 4186    & 4274    & 4318    & 4602    & 4924    \\
\bottomrule
\end{tabular}
\end{table*}

We analyzed the effect of the switch count cap ($C_{max}$) on accuracy and generation length using Qwen3-1.7B on MATH500 in Tab.~\ref{tab:cmax}. The results indicate that
\begin{itemize}[leftmargin=*,labelsep=0.5em]
  \item Efficiency control: As $C_{max}$ decreases (from $\infty$ down to 4), the generation length drops significantly (from 4924 to 1189), demonstrating the controller's effectiveness in curbing overthinking. This provides users with sufficient capabilities to control the intensity of thinking flexibly.
  \item Accuracy trade-off: While extremely tight caps (\textit{e.g.}, 4 or 8) limit accuracy, moderate caps (\textit{e.g.}, 20-32) achieve performance competitive with or close to the unlimited setting ($C_{max}=\infty$) but with much lower token consumption. This confirms that the controller effectively makes use of partial reasoning trajectories to deliver reasonable answers.
\end{itemize}

\subsection{Detailed Evaluation Results under Varying Token Budgets}
\label{supp_exp_eff}

We provide detailed evaluation results of Qwen3-8B in Tab.~\ref{tab:gsm8k_len_sorted}-\ref{tab:aime25_len_sorted}, Qwen3-1.7B in Tab.~\ref{tab:gsm8k_qwen1p7b_len_sorted}-\ref{tab:aime25_qwen1p7b_len_sorted}, and DeepSeek-R1-Distill-Llama-8B in Tab.~\ref{tab:gsm8k_deepseek_len_sorted}-\ref{tab:aime25_deepseek_len_sorted}.

\begin{table*}[ht]
\scriptsize
\centering
\captionsetup{font={small}}
\setlength{\tabcolsep}{6pt}
\renewcommand{\arraystretch}{1.15}
\caption{Evaluation results of Qwen3-8B on the GSM8K benchmark under varying token budgets. Rows are \textit{sorted by generation length in descending order}.}
\label{tab:gsm8k_len_sorted}

\begin{BarScale}{96.06}{2240}
\begin{tabular}{lcc}
\toprule
\makecell[l]{\textbf{Method}} & \makecell[c]{\textbf{Accuracy (\%)}} & \makecell[c]{\textbf{Generation Length}} \\
\midrule
CoT (Greedy)                          & \accbarv{95.68}\, 95.68\% & \genbarv{2240}\, 2240\\
\rowcolor{oliveGreen!30}\textbf{\textcolor{oliveGreen}{SwiR (Ours)}} & \accbarv{96.06}\, 96.06\% & \genbarv{2218}\, 2218\\
CoT (Greedy)                          & \accbarv{95.75}\, 95.75\% & \genbarv{2199}\, 2199\\
CoT                                   & \accbarv{95.60}\, 95.60\% & \genbarv{2138}\, 2138\\
CoT                                   & \accbarv{95.60}\, 95.60\% & \genbarv{2136}\, 2136\\
CoT                                   & \accbarv{94.77}\, 94.77\% & \genbarv{2123}\, 2123\\
CoT (Greedy)                          & \accbarv{95.15}\, 95.15\% & \genbarv{2115}\, 2115\\
\bottomrule
\end{tabular}
\end{BarScale}
\end{table*}

\begin{table*}[ht]
\scriptsize
\centering
\captionsetup{font={small}}
\setlength{\tabcolsep}{6pt}
\renewcommand{\arraystretch}{1.15}
\caption{Evaluation results of Qwen3-8B on the MATH500 benchmark under varying token budgets. Rows are \textit{sorted by generation length in descending order}.}
\label{tab:math500_len_sorted}

\begin{BarScale}{98.40}{5311}
\begin{tabular}{lcc}
\toprule
\makecell[l]{\textbf{Method}} & \makecell[c]{\textbf{Accuracy (\%)}} & \makecell[c]{\textbf{Generation Length}} \\
\midrule
CoT (Greedy)                                                       & \accbarv{96.40}\, 96.40\% & \genbarv{5311}\, 5311\\
\rowcolor{oliveGreen!30}\textbf{\textcolor{oliveGreen}{SwiR (Ours)}} & \accbarv{98.40}\, 98.40\% & \genbarv{5183}\, 5183\\
CoT                                                                & \accbarv{96.00}\, 96.00\% & \genbarv{4985}\, 4985\\
Soft Thinking                                                      & \accbarv{96.00}\, 96.00\% & \genbarv{4934}\, 4934\\
Soft Thinking                                                      & \accbarv{95.40}\, 95.40\% & \genbarv{4733}\, 4733\\
CoT                                                                & \accbarv{95.60}\, 95.60\% & \genbarv{4729}\, 4729\\
CoT (Greedy)                                                       & \accbarv{94.00}\, 94.00\% & \genbarv{4565}\, 4565\\
\bottomrule
\end{tabular}
\end{BarScale}
\end{table*}

\begin{table*}[ht]
\scriptsize
\centering
\captionsetup{font={small}}
\setlength{\tabcolsep}{6pt}
\renewcommand{\arraystretch}{1.15}
\caption{Evaluation results of Qwen3-8B on the GPQA Diamond benchmark under varying token budgets. Rows are \textit{sorted by generation length in descending order}.}
\label{tab:gpqa_len_sorted}

\begin{BarScale}{61.11}{8359}
\begin{tabular}{lcc}
\toprule
\makecell[l]{\textbf{Method}} & \makecell[c]{\textbf{Accuracy (\%)}} & \makecell[c]{\textbf{Generation Length}} \\
\midrule
\rowcolor{oliveGreen!30}\textbf{\textcolor{oliveGreen}{SwiR (Ours)}} & \accbarv{61.11}\, 61.11\% & \genbarv{8359}\, 8359\\
Soft Thinking                                                      & \accbarv{59.59}\, 59.59\% & \genbarv{8153}\, 8153\\
CoT                                                               & \accbarv{59.60}\, 59.60\% & \genbarv{8123}\, 8123\\
CoT (Greedy)                                                      & \accbarv{56.57}\, 56.57\% & \genbarv{7909}\, 7909\\
CoT                                                               & \accbarv{55.56}\, 55.56\% & \genbarv{7570}\, 7570\\
CoT (Greedy)                                                      & \accbarv{55.05}\, 55.05\% & \genbarv{7546}\, 7546\\
Soft Thinking                                                     & \accbarv{55.05}\, 55.05\% & \genbarv{7433}\, 7433\\
\bottomrule
\end{tabular}
\end{BarScale}
\end{table*}

\begin{table*}[ht]
\scriptsize
\centering
\captionsetup{font={small}}
\setlength{\tabcolsep}{6pt}
\renewcommand{\arraystretch}{1.15}
\caption{Evaluation results of Qwen3-8B on the AIME2024 benchmark under varying token budgets. Rows are \textit{sorted by generation length in descending order}.}
\label{tab:aime24_len_sorted}

\begin{BarScale}{79.17}{12491}
\begin{tabular}{lcc}
\toprule
\makecell[l]{\textbf{Method}} & \makecell[c]{\textbf{Accuracy (\%)}} & \makecell[c]{\textbf{Generation Length}} \\
\midrule
\rowcolor{oliveGreen!30}\textbf{\textcolor{oliveGreen}{SwiR (Ours)}} & \accbarv{79.17}\, 79.17\% & \genbarv{12491}\, 12491\\
Soft Thinking                                                      & \accbarv{67.92}\, 67.92\% & \genbarv{12271}\, 12271\\
CoT                                                               & \accbarv{75.83}\, 75.83\% & \genbarv{12077}\, 12077\\
CoT (Greedy)                                                      & \accbarv{70.00}\, 70.00\% & \genbarv{11680}\, 11680\\
\rowcolor{oliveGreen!30}\textbf{\textcolor{oliveGreen}{SwiR (Ours)}} & \accbarv{69.58}\, 69.58\% & \genbarv{10815}\, 10815\\
\rowcolor{oliveGreen!30}\textbf{\textcolor{oliveGreen}{SwiR (Ours)}} & \accbarv{66.25}\, 66.25\% & \genbarv{10349}\, 10349\\
CoT (Greedy)                                                      & \accbarv{66.67}\, 66.67\% & \genbarv{10328}\, 10328\\
\bottomrule
\end{tabular}
\end{BarScale}
\end{table*}

\begin{table*}[ht]
\scriptsize
\centering
\captionsetup{font={small}}
\setlength{\tabcolsep}{6pt}
\renewcommand{\arraystretch}{1.15}
\caption{Evaluation results of Qwen3-8B on the AIME2025 benchmark under varying token budgets. Rows are \textit{sorted by generation length in descending order}.}
\label{tab:aime25_len_sorted}

\begin{BarScale}{70.00}{13911}
\begin{tabular}{lcc}
\toprule
\makecell[l]{\textbf{Method}} & \makecell[c]{\textbf{Accuracy (\%)}} & \makecell[c]{\textbf{Generation Length}} \\
\midrule
\rowcolor{oliveGreen!30}\textbf{\textcolor{oliveGreen}{SwiR (Ours)}} & \accbarv{70.00}\, 70.00\% & \genbarv{13911}\, 13911\\
Soft Thinking                                                      & \accbarv{68.33}\, 68.33\% & \genbarv{13665}\, 13665\\
CoT (Greedy)                                                      & \accbarv{60.00}\, 60.00\% & \genbarv{13292}\, 13292\\
CoT                                                               & \accbarv{67.50}\, 67.50\% & \genbarv{12924}\, 12924\\
\rowcolor{oliveGreen!30}\textbf{\textcolor{oliveGreen}{SwiR (Ours)}} & \accbarv{62.92}\, 62.92\% & \genbarv{11482}\, 11482\\
\rowcolor{oliveGreen!30}\textbf{\textcolor{oliveGreen}{SwiR (Ours)}} & \accbarv{58.33}\, 58.33\% & \genbarv{10596}\, 10596\\
CoT                                                               & \accbarv{54.17}\, 54.17\% & \genbarv{10215}\, 10215\\
\bottomrule
\end{tabular}
\end{BarScale}
\end{table*}

\begin{table*}[ht]
\scriptsize
\centering
\captionsetup{font={small}}
\setlength{\tabcolsep}{6pt}
\renewcommand{\arraystretch}{1.15}
\caption{Evaluation results of Qwen3-1.7B on the GSM8K benchmark under varying token budgets. Rows are \textit{sorted by generation length in descending order}.}
\label{tab:gsm8k_qwen1p7b_len_sorted}

\begin{BarScale}{90.83}{2038}
\begin{tabular}{lcc}
\toprule
\makecell[l]{\textbf{Method}} & \makecell[c]{\textbf{Accuracy (\%)}} & \makecell[c]{\textbf{Generation Length}} \\
\midrule
CoT (Greedy)                                                        & \accbarv{89.61}\, 89.61\% & \genbarv{2038}\, 2038\\
\rowcolor{oliveGreen!30}\textbf{\textcolor{oliveGreen}{SwiR (Ours)}} & \accbarv{90.83}\, 90.83\% & \genbarv{2010}\, 2010\\
CoT                                                                 & \accbarv{90.45}\, 90.45\% & \genbarv{1981}\, 1981\\
CoT (Greedy)                                                        & \accbarv{89.61}\, 89.61\% & \genbarv{1968}\, 1968\\
Soft Thinking                                                       & \accbarv{90.30}\, 90.30\% & \genbarv{1959}\, 1959\\
Soft Thinking                                                       & \accbarv{90.22}\, 90.22\% & \genbarv{1946}\, 1946\\
CoT                                                                 & \accbarv{89.23}\, 89.23\% & \genbarv{1928}\, 1928\\
\bottomrule
\end{tabular}
\end{BarScale}
\end{table*}

\begin{table*}[ht]
\scriptsize
\centering
\captionsetup{font={small}}
\setlength{\tabcolsep}{6pt}
\renewcommand{\arraystretch}{1.15}
\caption{Evaluation results of Qwen3-1.7B on the MATH500 benchmark under varying token budgets. Rows are \textit{sorted by generation length in descending order}.}
\label{tab:math500_qwen1p7b_len_sorted}

\begin{BarScale}{93.00}{4924}
\begin{tabular}{lcc}
\toprule
\makecell[l]{\textbf{Method}} & \makecell[c]{\textbf{Accuracy (\%)}} & \makecell[c]{\textbf{Generation Length}} \\
\midrule
\rowcolor{oliveGreen!30}\textbf{\textcolor{oliveGreen}{SwiR (Ours)}} & \accbarv{93.00}\, 93.00\% & \genbarv{4924}\, 4924\\
CoT (Greedy)                                                      & \accbarv{91.00}\, 91.00\% & \genbarv{4799}\, 4799\\
CoT                                                               & \accbarv{92.00}\, 92.00\% & \genbarv{4780}\, 4780\\
Soft Thinking                                                     & \accbarv{90.60}\, 90.60\% & \genbarv{4721}\, 4721\\
CoT                                                               & \accbarv{90.80}\, 90.80\% & \genbarv{4435}\, 4435\\
CoT (Greedy)                                                      & \accbarv{89.20}\, 89.20\% & \genbarv{4342}\, 4342\\
Soft Thinking                                                     & \accbarv{89.00}\, 89.00\% & \genbarv{4288}\, 4288\\
\bottomrule
\end{tabular}
\end{BarScale}
\end{table*}

\begin{table*}[ht]
\scriptsize
\centering
\captionsetup{font={small}}
\setlength{\tabcolsep}{6pt}
\renewcommand{\arraystretch}{1.15}
\caption{Evaluation results of Qwen3-1.7B on the GPQA Diamond benchmark under varying token budgets. Rows are \textit{sorted by generation length in descending order}.}
\label{tab:gpqa_qwen1p7b_len_sorted}

\begin{BarScale}{41.41}{9517}
\begin{tabular}{lcc}
\toprule
\makecell[l]{\textbf{Method}} & \makecell[c]{\textbf{Accuracy (\%)}} & \makecell[c]{\textbf{Generation Length}} \\
\midrule
\rowcolor{oliveGreen!30}\textbf{\textcolor{oliveGreen}{SwiR (Ours)}} & \accbarv{41.41}\, 41.41\% & \genbarv{9517}\, 9517\\
CoT (Greedy)                                                      & \accbarv{31.82}\, 31.82\% & \genbarv{9190}\, 9190\\
CoT (Greedy)                                                      & \accbarv{30.30}\, 30.30\% & \genbarv{8751}\, 8751\\
Soft Thinking                                                     & \accbarv{34.34}\, 34.34\% & \genbarv{8731}\, 8731\\
CoT                                                               & \accbarv{39.39}\, 39.39\% & \genbarv{8625}\, 8625\\
CoT                                                               & \accbarv{37.37}\, 37.37\% & \genbarv{8040}\, 8040\\
\rowcolor{oliveGreen!30}\textbf{\textcolor{oliveGreen}{SwiR (Ours)}} & \accbarv{36.87}\, 36.87\% & \genbarv{7773}\, 7773\\
\bottomrule
\end{tabular}
\end{BarScale}
\end{table*}

\begin{table*}[ht]
\scriptsize
\centering
\captionsetup{font={small}}
\setlength{\tabcolsep}{6pt}
\renewcommand{\arraystretch}{1.15}
\caption{Evaluation results of Qwen3-1.7B on the AIME2024 benchmark under varying token budgets. Rows are \textit{sorted by generation length in descending order}.}
\label{tab:aime24_qwen1p7b_len_sorted}

\begin{BarScale}{50.83}{12825}
\begin{tabular}{lcc}
\toprule
\makecell[l]{\textbf{Method}} & \makecell[c]{\textbf{Accuracy (\%)}} & \makecell[c]{\textbf{Generation Length}} \\
\midrule
CoT (Greedy)                                                        & \accbarv{40.00}\, 40.00\% & \genbarv{12825}\, 12825\\
\rowcolor{oliveGreen!30}\textbf{\textcolor{oliveGreen}{SwiR (Ours)}} & \accbarv{50.83}\, 50.83\% & \genbarv{12702}\, 12702\\
CoT                                                                 & \accbarv{45.83}\, 45.83\% & \genbarv{11896}\, 11896\\
Soft Thinking                                                       & \accbarv{38.75}\, 38.75\% & \genbarv{10788}\, 10788\\
\rowcolor{oliveGreen!30}\textbf{\textcolor{oliveGreen}{SwiR (Ours)}} & \accbarv{42.08}\, 42.08\% & \genbarv{10243}\, 10243\\
Soft Thinking                                                       & \accbarv{36.67}\, 36.67\% & \genbarv{9841}\, 9841\\
CoT                                                                 & \accbarv{38.33}\, 38.33\% & \genbarv{9510}\, 9510\\
\bottomrule
\end{tabular}
\end{BarScale}
\end{table*}

\begin{table*}[ht]
\scriptsize
\centering
\captionsetup{font={small}}
\setlength{\tabcolsep}{6pt}
\renewcommand{\arraystretch}{1.15}
\caption{Evaluation results of Qwen3-1.7B on the AIME2025 benchmark under varying token budgets. Rows are \textit{sorted by generation length in descending order}.}
\label{tab:aime25_qwen1p7b_len_sorted}

\begin{BarScale}{38.33}{11408}
\begin{tabular}{lcc}
\toprule
\makecell[l]{\textbf{Method}} & \makecell[c]{\textbf{Accuracy (\%)}} & \makecell[c]{\textbf{Generation Length}} \\
\midrule
CoT (Greedy)                                                      & \accbarv{33.33}\, 33.33\% & \genbarv{11408}\, 11408\\
\rowcolor{oliveGreen!30}\textbf{\textcolor{oliveGreen}{SwiR (Ours)}} & \accbarv{38.33}\, 38.33\% & \genbarv{9944}\, 9944\\
CoT                                                               & \accbarv{33.33}\, 33.33\% & \genbarv{9733}\, 9733\\
Soft Thinking                                                     & \accbarv{36.67}\, 36.67\% & \genbarv{8904}\, 8904\\
CoT (Greedy)                                                      & \accbarv{26.67}\, 26.67\% & \genbarv{8890}\, 8890\\
CoT                                                               & \accbarv{31.25}\, 31.25\% & \genbarv{8618}\, 8618\\
\rowcolor{oliveGreen!30}\textbf{\textcolor{oliveGreen}{SwiR (Ours)}} & \accbarv{34.58}\, 34.58\% & \genbarv{8543}\, 8543\\
\bottomrule
\end{tabular}
\end{BarScale}
\end{table*}

\begin{table*}[ht]
\scriptsize
\centering
\captionsetup{font={small}}
\setlength{\tabcolsep}{6pt}
\renewcommand{\arraystretch}{1.15}
\caption{Evaluation results of DeepSeek-R1-Distill-Llama-8B on the GSM8K benchmark under varying token budgets. Rows are \textit{sorted by generation length in descending order}.}
\label{tab:gsm8k_deepseek_len_sorted}

\begin{BarScale}{90.07}{2953}
\begin{tabular}{lcc}
\toprule
\makecell[l]{\textbf{Method}} & \makecell[c]{\textbf{Accuracy (\%)}} & \makecell[c]{\textbf{Generation Length}} \\
\midrule
Soft Thinking                                                      & \accbarv{85.90}\, 85.90\% & \genbarv{2953}\, 2953\\
Soft Thinking                                                      & \accbarv{84.84}\, 84.84\% & \genbarv{2516}\, 2516\\
CoT (Greedy)                                                       & \accbarv{85.82}\, 85.82\% & \genbarv{2393}\, 2393\\
Soft Thinking                                                      & \accbarv{83.62}\, 83.62\% & \genbarv{2266}\, 2266\\
Soft Thinking                                                      & \accbarv{77.41}\, 77.41\% & \genbarv{1741}\, 1741\\
CoT (Greedy)                                                       & \accbarv{86.05}\, 86.05\% & \genbarv{1642}\, 1642\\
CoT                                                                & \accbarv{89.46}\, 89.46\% & \genbarv{1588}\, 1588\\
\bottomrule
\end{tabular}
\end{BarScale}
\end{table*}

\begin{table*}[ht]
\scriptsize
\centering
\captionsetup{font={small}}
\setlength{\tabcolsep}{6pt}
\renewcommand{\arraystretch}{1.15}
\caption{Evaluation results of DeepSeek-R1-Distill-Llama-8B on the Math500 benchmark under varying token budgets. Rows are \textit{sorted by generation length in descending order}.}
\label{tab:math500_deepseek_len_sorted}

\begin{BarScale}{92.00}{4718}
\begin{tabular}{lcc}
\toprule
\makecell[l]{\textbf{Method}} & \makecell[c]{\textbf{Accuracy (\%)}} & \makecell[c]{\textbf{Generation Length}} \\
\midrule
Soft Thinking                                                      & \accbarv{83.80}\, 83.80\% & \genbarv{4718}\, 4718\\
CoT (Greedy)                                                       & \accbarv{84.80}\, 84.80\% & \genbarv{4110}\, 4110\\
Soft Thinking                                                      & \accbarv{82.60}\, 82.60\% & \genbarv{4085}\, 4085\\
\rowcolor{oliveGreen!30}\textbf{\textcolor{oliveGreen}{SwiR (Ours)}} & \accbarv{92.00}\, 92.00\% & \genbarv{3837}\, 3837\\
CoT                                                                & \accbarv{91.40}\, 91.40\% & \genbarv{3792}\, 3792\\
CoT                                                                & \accbarv{89.80}\, 89.80\% & \genbarv{3572}\, 3572\\
Soft Thinking                                                      & \accbarv{75.40}\, 75.40\% & \genbarv{3204}\, 3204\\
\bottomrule
\end{tabular}
\end{BarScale}
\end{table*}

\begin{table*}[ht]
\scriptsize
\centering
\captionsetup{font={small}}
\setlength{\tabcolsep}{6pt}
\renewcommand{\arraystretch}{1.15}
\caption{Evaluation results of DeepSeek-R1-Distill-Llama-8B on the GPQA Diamond benchmark under varying token budgets. Rows are \textit{sorted by generation length in descending order}.}
\label{tab:gpqa_deepseek_len_sorted}

\begin{BarScale}{47.98}{8593}
\begin{tabular}{lcc}
\toprule
\makecell[l]{\textbf{Method}} & \makecell[c]{\textbf{Accuracy (\%)}} & \makecell[c]{\textbf{Generation Length}} \\
\midrule
Soft Thinking                                                      & \accbarv{33.33}\, 33.33\% & \genbarv{8593}\, 8593\\
CoT                                                                & \accbarv{46.46}\, 46.46\% & \genbarv{7591}\, 7591\\
Soft Thinking                                                      & \accbarv{30.81}\, 30.81\% & \genbarv{7507}\, 7507\\
\rowcolor{oliveGreen!30}\textbf{\textcolor{oliveGreen}{SwiR (Ours)}} & \accbarv{47.98}\, 47.98\% & \genbarv{7458}\, 7458\\
CoT                                                                & \accbarv{46.46}\, 46.46\% & \genbarv{7236}\, 7236\\
\rowcolor{oliveGreen!30}\textbf{\textcolor{oliveGreen}{SwiR (Ours)}} & \accbarv{45.45}\, 45.45\% & \genbarv{6635}\, 6635\\
\rowcolor{oliveGreen!30}\textbf{\textcolor{oliveGreen}{SwiR (Ours)}} & \accbarv{44.44}\, 44.44\% & \genbarv{6038}\, 6038\\
\bottomrule
\end{tabular}
\end{BarScale}
\end{table*}

\begin{table*}[ht]
\scriptsize
\centering
\captionsetup{font={small}}
\setlength{\tabcolsep}{6pt}
\renewcommand{\arraystretch}{1.15}
\caption{Evaluation results of DeepSeek-R1-Distill-Llama-8B on the AIME2024 benchmark under varying token budgets. Rows are \textit{sorted by generation length in descending order}.}
\label{tab:aime24_deepseek_len_sorted_v2}

\begin{BarScale}{45.00}{8179}
\begin{tabular}{lcc}
\toprule
\makecell[l]{\textbf{Method}} & \makecell[c]{\textbf{Accuracy (\%)}} & \makecell[c]{\textbf{Generation Length}} \\
\midrule
\rowcolor{oliveGreen!30}\textbf{\textcolor{oliveGreen}{SwiR (Ours)}} & \accbarv{45.00}\, 45.00\% & \genbarv{8179}\, 8179\\
CoT                                                                 & \accbarv{43.75}\, 43.75\% & \genbarv{8145}\, 8145\\
CoT (Greedy)                                                        & \accbarv{30.00}\, 30.00\% & \genbarv{7840}\, 7840\\
\rowcolor{oliveGreen!30}\textbf{\textcolor{oliveGreen}{SwiR (Ours)}} & \accbarv{44.17}\, 44.17\% & \genbarv{7555}\, 7555\\
CoT                                                                 & \accbarv{41.25}\, 41.25\% & \genbarv{7330}\, 7330\\
CoT (Greedy)                                                        & \accbarv{26.66}\, 26.66\% & \genbarv{7107}\, 7107\\
Soft Thinking                                                       & \accbarv{34.17}\, 34.17\% & \genbarv{6956}\, 6956\\
\bottomrule
\end{tabular}
\end{BarScale}
\end{table*}

\begin{table*}[ht]
\scriptsize
\centering
\captionsetup{font={small}}
\setlength{\tabcolsep}{6pt}
\renewcommand{\arraystretch}{1.15}
\caption{Evaluation results of DeepSeek-R1-Distill-Llama-8B on the AIME2025 benchmark under varying token budgets. Rows are \textit{sorted by generation length in descending order}.}
\label{tab:aime25_deepseek_len_sorted}

\begin{BarScale}{31.25}{8448}
\begin{tabular}{lcc}
\toprule
\makecell[l]{\textbf{Method}} & \makecell[c]{\textbf{Accuracy (\%)}} & \makecell[c]{\textbf{Generation Length}} \\
\midrule
Soft Thinking                                                      & \accbarv{20.42}\, 20.42\% & \genbarv{8448}\, 8448\\
\rowcolor{oliveGreen!30}\textbf{\textcolor{oliveGreen}{SwiR (Ours)}} & \accbarv{31.25}\, 31.25\% & \genbarv{6827}\, 6827\\
Soft Thinking                                                      & \accbarv{19.17}\, 19.17\% & \genbarv{6824}\, 6824\\
CoT                                                                & \accbarv{26.25}\, 26.25\% & \genbarv{6583}\, 6583\\
\rowcolor{oliveGreen!30}\textbf{\textcolor{oliveGreen}{SwiR (Ours)}} & \accbarv{30.83}\, 30.83\% & \genbarv{6419}\, 6419\\
CoT (Greedy)                                                       & \accbarv{30.00}\, 30.00\% & \genbarv{6293}\, 6293\\
\rowcolor{oliveGreen!30}\textbf{\textcolor{oliveGreen}{SwiR (Ours)}} & \accbarv{30.00}\, 30.00\% & \genbarv{6230}\, 6230\\
\bottomrule
\end{tabular}
\end{BarScale}
\end{table*}

\end{document}